\documentclass[lettersize,journal]{IEEEtran}
\IEEEoverridecommandlockouts
\usepackage{cite}
\usepackage{comment}
\usepackage{amsmath,amssymb,amsfonts}
\usepackage{algorithm}
\usepackage{algorithmic}
\usepackage{graphicx}
\usepackage{textcomp}
\usepackage{tabularx}
\newenvironment{tableitemize}
{ \begin{minipage}[t]{\linewidth} \vspace{-10pt} \hspace{5pt} \begin{itemize}[leftmargin=10pt] \vspace{5pt}}
{  \vspace{5pt} \end{itemize} \end{minipage}   } 
\usepackage{pifont}
\newcommand{\cmark}{\textcolor{Green}{\ding{51}}}%
\newcommand{\xmark}{\textcolor{Red}{\ding{55}}}%
\newcommand{\tmark}{\textcolor{Goldenrod}{\ding{115}}}%
\usepackage[dvipsnames]{xcolor}
\usepackage{diagbox}
\usepackage{adjustbox}
\usepackage{multirow}
\usepackage[utf8]{inputenc}
\usepackage{booktabs, makecell}
\usepackage{dsfont}
\usepackage[caption=false,font=footnotesize,labelfont=sf,textfont=sf]{subfig}
\usepackage{hyperref}
\let\labelindent\relax
\usepackage[inline]{enumitem}

\usepackage{threeparttable}
\usepackage{blindtext}
\usepackage{geometry}
\def\BibTeX{{\rm B\kern-.05em{\sc i\kern-.025em b}\kern-.08em
    T\kern-.1667em\lower.7ex\hbox{E}\kern-.125emX}}

\begin{document}

\title{Federated Learning for Connected and Automated Vehicles: A Survey of Existing Approaches and Challenges}
\author{Vishnu Pandi Chellapandi,~\IEEEmembership{Member,~IEEE},
        Liangqi Yuan,~\IEEEmembership{Student Member,~IEEE},
        Christopher G. Brinton,~\IEEEmembership{Senior Member,~IEEE},
        Stanislaw H. \.{Z}ak,~\IEEEmembership{Life Member,~IEEE},
        and Ziran Wang,~\IEEEmembership{Member,~IEEE}
\thanks{V. P. Chellapandi, L. Yuan, C. G. Brinton, S. H. \.{Z}ak, and Z. Wang are with the College of Engineering, Purdue University, West Lafayette, IN 47907, USA (e-mails: cvp@purdue.edu, liangqiy@purdue.edu, cgb@purdue.edu, zak@purdue.edu, ryanwang11@hotmail.com).}}

\markboth{IEEE Transactions on Intelligent Vehicles, Early Access}%
{Chellapandi \MakeLowercase{\textit{et al.}}: Federated Learning for Connected and Automated Vehicles: A Survey of Existing Approaches and Challenges}

\maketitle
\begin{abstract}

Machine learning (ML) is widely used for key tasks in Connected and Automated Vehicles (CAV), including perception, planning, and control. However, its reliance on vehicular data for model training presents significant challenges related to in-vehicle user privacy and communication overhead generated by massive data volumes. Federated learning (FL) is a decentralized ML approach that enables multiple vehicles to collaboratively develop models, broadening learning from various driving environments, enhancing overall performance, and simultaneously securing local vehicle data privacy and security. This survey paper presents a review of the advancements made in the application of FL for CAV (FL4CAV). First, centralized and decentralized frameworks of FL are analyzed, highlighting their key characteristics and methodologies. Second, diverse data sources, models, and data security techniques relevant to FL in CAVs are reviewed, emphasizing their significance in ensuring privacy and confidentiality. Third, specific applications of FL are explored, providing insight into the base models and datasets employed for each application. \textcolor{black}{Finally, existing challenges for FL4CAV are listed and potential directions for future investigation to further enhance the effectiveness and efficiency of FL in the context of CAV are discussed.} \end{abstract}


\newcommand{\etal}{\textit{et al.}}

\section{Introduction}
\label{Sec. Introduction}
\IEEEPARstart{C}{onnected} and automated vehicles (CAV) are the key to future intelligent transportation systems (ITS) that encompass both ground and air transportation~\cite{pan2021flying,zhang2022intelligent,biparva2022video,cao2022future,hadjigeorgiou2022real,pei2022fault,guo2022hierarchical,wang2023new,zhang2022parallel}. With the advent of big data, the Internet of Things (IoT), edge computing, and intelligent systems, CAVs have the potential to improve the overall transportation system by reducing traffic accidents, congestion, and pollution~\cite{liu2021vision,chen2022milestones,hu2022distributed,tang2022prediction}. 
\textcolor{black}{CAVs integrate both Vehicle-to-Vehicle (V2V) and Vehicle-to-Infrastructure (V2I) communication capabilities, facilitating an enhanced perception of the environment beyond the direct line of sight~\cite{wang2018cluster,wang2021digital,wang2022conflict}. }This involves interaction with other vehicles, traffic signals, pedestrians, and other elements of the transportation ecosystem. Furthermore, CAVs are designed to assume control of driving tasks by the human operator under certain conditions, using a variety of sensors and sophisticated machine learning (ML) algorithms to achieve autonomous operation.

\textcolor{black}{Currently, CAVs are generating a tremendous amount of raw data, between 20 and 40 TB per day, per vehicle~\cite{cavdata2020}. The various sources of these data include engine components, electronic control units (ECU), perception sensors, and vehicle-to-everything (V2X) communications.} This large amount of data is sent to other vehicles, roadside infrastructures, or the cloud, continuously or periodically for monitoring, prognostics, diagnostics, and connectivity features~\cite{wang2022mobility}. \textcolor{black}{This flow of data has driven the flourishing deployment and application of ML in CAVs, including areas such as Advanced Driver-Assistance Systems (ADAS)~\cite{nidamanuri2021progressive}, automated driving~\cite{natan2022end}, ITS~\cite{park2023multi}, and sustainable development~\cite{singh2021deep}.}

\begin{figure*}[t]
\centering
\includegraphics[width=\linewidth]{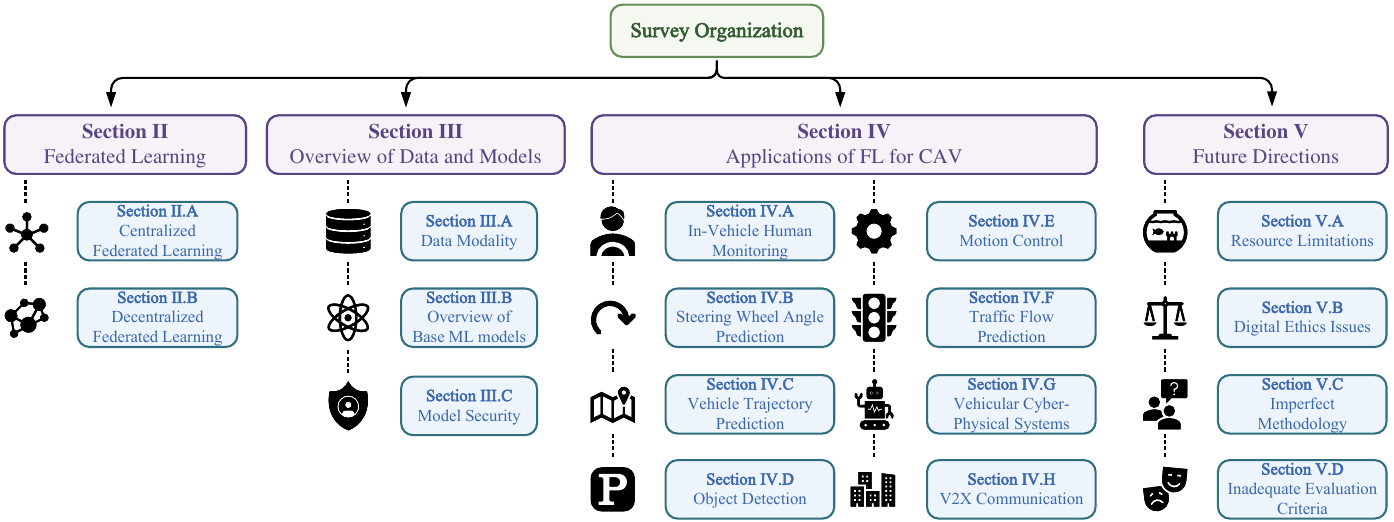}
\caption{Roadmap of this survey paper.}
\label{Fig. Roadmap}
\end{figure*}

\subsection{Motivation}

Due to the large amount of data required to train ML models, concerns have been raised about data security in terms of the legitimacy of data collection, data misuse, and privacy breaches. Data collected by various sensors in CAVs, are also considered private and are subject to stringent privacy protection regulations in different regions. One such example is the General Data Protection Regulation (GDPR) in the European Union \cite{voigt2017eu}, which imposes strict requirements and guidelines on the handling and processing of personal data to ensure individuals' privacy rights are protected. Even with the development of advanced ML techniques and vehicle connectivity, it has not been feasible to have a secure framework to collect data from every vehicle and train an ML model. These limitations led to the development of a new ML paradigm known as Federated Learning (FL) \cite{yang_federated_2019,kairouz_advances_2021}. 
\textcolor{black}{The term Federated Learning (FL) has been coined by Google~\cite{mcmahan_communication-efficient_2017}. FL was initially used for mobile keyboard prediction in Gboard~\cite{hard_federated_2019} to allow multiple mobile phones to cooperatively and securely train an ML model}. FL has been extensively applied in various fields such as industry \cite{zhang2020blockchain,zeng2022homophily,zhang2022r}, energy \cite{zhang2022semi,chen2023knowledge}, healthcare \cite{zhang2021dynamic, chen2023dfml}, and more.

In FL, edge devices/clients only send the gradients or the learnable parameters to cloud servers rather than sending massive local datasets in a centralized learning framework. Cloud servers perform a secure aggregation of the received gradients/weights and update the global model parameters that are transmitted back to clients/edge devices~\cite{bonawitz2017practical}. This procedure, known as a communication round, continues iteratively until the convergence criteria are met in the global model optimization. The key advantage of FL is reducing the strain on the network while also preserving the privacy of the local data. FL is a potential candidate that can utilize the data available from each CAV and develop a robust ML model. 

\textcolor{black}{Despite the benefits of V2X communications among CAVs, the invasion of privacy, accuracy, effectiveness, and communication resources is an essential concern to be addressed.} FL frameworks have received attention for their natural ability to preserve privacy by transmitting only model data between the server and its clients without including local vehicle data. In particular, the model data packets are smaller than the user data, thus saving the consumption of communication resources. Similarly, FL frameworks distribute training tasks to each client, and the server does not perform training but only aggregates, which can reduce the computational demand on the server and improves training efficiency. 
\textcolor{black}{Recently, there have also been efforts to train a decentralized FL that allows multiple vehicles to collaboratively train a model without needing a central server~\cite{nedic2018network,chellapandi2023convergence}.
In our first survey of FL for CAV (FL4CAV) presented in \cite{chellapandi2023survey}, we emphasized applications and explored foundational challenges in the subject.} Building upon that conference version, this extended journal paper further delves into the underlying methodologies, provides a more comprehensive review of recent developments, and introduces novel insights and evaluations, thereby presenting a more exhaustive and nuanced understanding of the field.

\begin{table*}[t]
\caption{Comparison of Related Surveys of Federated Learning for Connected and Automated Vehicles}
\label{Table Comparison of related survey}
\centering
\scriptsize
\begin{tabularx}{\linewidth}{|l|l|l|l|p{40pt}|p{35pt}|X|}
\hline
\textbf{Survey} & \textbf{Time} & \textbf{Focused topic in FL} & \textbf{CFL \& DFL}  & \textbf{Vehicle Data Modality} & \textbf{Vehicle \ \ Application} & \textbf{Highlights} \\
\hline
Du \textit{et al.} \cite{du2020federated} & 2020 & Vehicular IoT & \xmark & \xmark & \xmark & \begin{tableitemize} \item First survey of FL in vehicular IoT \end{tableitemize} \\
\hline
Jiang \textit{et al.} \cite{jiang2020federated} & 2020 & Smart city & \xmark & \xmark & \xmark & \begin{tableitemize} \item Opportunities of FL in the context of smart cities, such as interactions between CAVs and an urban sensing system. \end{tableitemize} \\
\hline
Savazzi \textit{et al.} \cite{savazzi2021opportunities} & 2021 & Automated industrial & \cmark & \xmark & \xmark & \begin{tableitemize} \item Opportunities of FL in next-generation connected industrial systems, including robotics, vehicles, and drones \end{tableitemize} \\
\hline
Nguyen \textit{et al.} \cite{nguyen2021federated} & 2021 & IoT & \cmark & \xmark & \xmark & \begin{tableitemize} \item FL in IoT applications, such as intelligent healthcare, transportation, city, unmanned aerial vehicles (UAV), and industrial \end{tableitemize} \\
\hline
Javed \textit{et al.} \cite{javed2022integration} & 2022 & Vehicular IoT Network & \xmark & \xmark & \xmark & \begin{tableitemize} \item Integrating blockchain and FL for vehicular IoT network. \end{tableitemize} \\
\hline

Yuan \textit{et al.} \cite{yuan2023decentralized} & 2023 & Decentralized FL & \cmark & \xmark & \xmark & \begin{tableitemize} \item Taxonomies and variants of DFL \item Analysis and state-of-the-art developments in different network topologies for DFL \end{tableitemize} \\
\hline

\textbf{This paper} & \textbf{2023} & \textbf{FL for CAV} & \cmark & \cmark & \cmark & \begin{tableitemize} \item \textbf{Advantages and disadvantages of CFL and DFL in CAV and state-of-the-art deployments.} \item \textbf{Diverse data modalities and security in CAV} \item \textbf{Facilitating 8 vehicle applications through FL} \item \textbf{Challenges and future research directions of FL for CAV across 4 major categories and 11 subcategories.} \end{tableitemize} \vspace{-5pt} \\
\hline
\end{tabularx}
\\[2pt] 
\cmark {} Yes, \xmark {} No.
\end{table*}

\begin{table*}[t]
\caption{Comparison of Machine Learning Approaches in Connected and Automated Vehicles}
\label{Table_Comparison_of_ML_Approaches}
\centering
\scriptsize
\begin{tabularx}{\linewidth}{|l|p{70pt}|p{70pt}|X|X|}
\hline
\textbf{Features} & \textbf{Edge Learning \newline (On-Vehicle only)} & \textbf{Centralized Learning \newline (On-Server only)} & \textbf{Centralized \newline Federated Learning} & \textbf{Decentralized \newline Federated Learning} \\
\hline
Model training & Local vehicle & Central server & Local vehicle training and central server aggregation & Local vehicle training and aggregation \\
\hline
Model applicability & Personalized model & Single global model & Single global model but can be personalized & Global models and personalized models \\
\hline
Privacy protection & \cmark\cmark & \xmark & \cmark & \tmark \\
\hline
Learning efficiency & \tmark & \cmark & \cmark\cmark & \cmark \\
\hline
Performance on heterogeneous/anomaly data & \tmark & \cmark\cmark & \cmark  & \cmark\cmark  \\
\hline
Communication (Data transmission) requirement & \cmark\cmark & \xmark & \tmark & \xmark \\
\hline
Training data volume & \xmark & \cmark\cmark & \cmark & \cmark \\
\hline
Current research progress & \cmark\cmark & \cmark\cmark & \tmark & \xmark \\
\hline
Compatibility with CAV & \cmark & \xmark & \cmark & \cmark\cmark \\
\hline
\end{tabularx}
\\[2pt] 
\cmark\cmark {} Very high, \cmark {} high, \tmark {} average, \xmark {} low.
\end{table*}

\subsection{Paper Organization}
In this paper, we provide a survey of FL4CAV, including deployment of various FL frameworks on CAVs, data modalities and security, diverse applications, and key challenges. The organization of this survey is shown in Fig.~\ref{Fig. Roadmap}. The following topics are covered in this survey:
\begin{itemize}
    \item A systematic review of FL algorithms is conducted, specifically focusing on their deployment in CAVs. Additionally, we examine the integration of ML models within the FL framework for CAV applications.
    \item Data modalities and data security considerations in CAVs are summarized, highlighting the diverse range of multi-modal data generated by various sensors.
    \item Critical applications of FL4CAV are explored, such as driver monitoring, steering wheel angle prediction, vehicle trajectory prediction, object detection, motion control application, traffic flow prediction, and V2X communications.
    \item Current challenges and future research directions of FL4CAV are highlighted, such as performance, safety, fairness, applicability, and scalability. 
    A comparison of our survey with other related surveys can be found in Table~\ref{Table Comparison of related survey}.
\end{itemize}

The remainder of this survey is organized as follows. \textcolor{black}{In Section~\ref{Sec.Algorithms}, we describe the two main FL frameworks along with their algorithms.} In Section~\ref{Sec. Overview}, we discuss various data modalities, ML methods used in FL4CAV applications, and FL data security in CAVs. \textcolor{black}{Section~\ref{Sec. Applications of FL for CAV} reviews various applications of FL in CAVs. The multi-modal data, algorithms, and datasets used in the relevant literature are also summarized. Challenges and potential research areas are discussed in Section~\ref{Sec. Challenges}. In Section~\ref{Sec. Conclusion}, we present conclusions of this survey.}

\begin{figure*}[ht]
\centering
\includegraphics[width=1\linewidth]{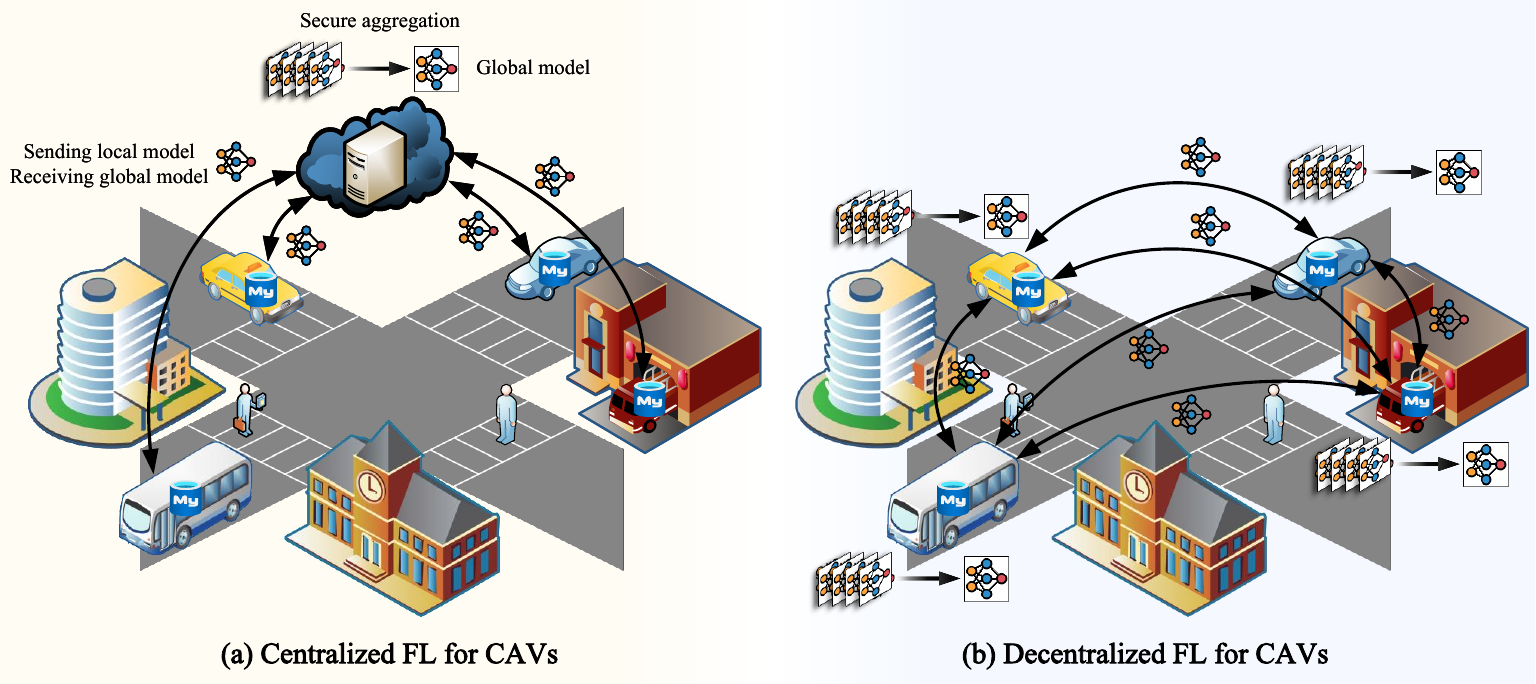}
\caption{Illustration of (a) centralized and (b) decentralized federated learning for connected and automated vehicles.}
\label{Fig. FL_CAV} 
\end{figure*}

\section{Federated Learning Methods}
\label{Sec.Algorithms}
In this section, we describe the FL frameworks in terms of two categories: centralized FL and decentralized FL. \textcolor{black}{An illustration of the categories is shown in Fig.~\ref{Fig. FL_CAV}. 
In addition, we provide an overview of the ML techniques that are commonly used as base models on local devices during the FL process. The steps of this process can be described as:}
\begin{enumerate}
\item \textit{Global Model Distribution}: The edge server disseminates the global model parameters to $K$ vehicles.
\item \textit{Model Update Using Local Data}: Each vehicle independently trains the ML model using its own local data. This training process typically adopts a simple Stochastic Gradient Descent (SGD) algorithm. The computational infrastructure is usually limited.
\item \textit{Local Update Upload}: After training the model, each vehicle applies privacy-preserving techniques such as differential privacy (introduces artificial noise to the parameters) and then uploads/communicates the model parameters to the selected central server (Centralized Federated Learning, i.e., CFL) or other vehicles (Decentralized Federated Learning, i.e., DFL).
\item \textit{Aggregation of Vehicle Updates}: The server securely aggregates the parameters uploaded from $K$ vehicles to obtain the global model. Furthermore, it tests the model's performance.
\end{enumerate}

\subsection{Centralized Federated Learning}
\textcolor{black}{In this section, we review two major aggregation methods in the centralized framework, namely averaging and a more recent technique called knowledge distillation.}

\subsubsection{Averaging}

Most of the existing literature uses the Federated Averaging (FedAvg) algorithm~\cite{mcmahan_communication-efficient_2017} for the FL aggregation process on the server---see Table~\ref{CFL Table Related literature}. FedAvg applies SGD optimization to local vehicles and performs a weighted averaging of the weights of the vehicles on the central server. FedAvg performs multiple local gradient updates before sending the parameters to the server, reducing the number of communication rounds.  For FL4CAV, data on each CAV are dynamically updated at each communication round.

\textcolor{black}{}{A typical FL setup has $K$ vehicles that have their own local data sets and the ability to perform simple local optimization. At the central server, the optimization problem can be represented as} 
\begin{equation}
    \label{eq:Objective-function}
    \min_{x\in \mathbb{R}^{d}}\Big[f(x)= \frac{1}{K}\sum_{i=1}^{K} f_i(x_i) \Big],
\end{equation}
where $f_i:\mathbb{R}^{d} \to\mathbb{R}$ for $i \in \{1, \dots , K\}$ is the local objective function of the $i^{th}$ vehicle. The local objective function of the $i^{th}$ vehicle can have the form,
\begin{equation}
    \label{eq:local-equation}
    f_i(x_i) = \mathbb{E}_{\xi_i \sim \mathcal{D}_{i}}[\ell(x_i,\xi_i)],
\end{equation}
where $\xi_i$ represents the data that have been sampled from the local vehicle data $\mathcal{D}_{i}$ for the $i^{th}$ vehicle. The expectation operator, $\mathbb{E}$, is acting on the local objective function, $\ell(x_i,\xi_i)$, with respect to a data sample, $\xi_i$, drawn from the vehicle data, $\mathcal{D}_{i}$. The function $\ell(x_i,\xi_i)$ is the loss function evaluated for each vehicle, $x_i$, and data sample, $\xi_i$. Here, $x_i \in \mathbb{R}^d$ represents the model parameters of vehicle $i$, and $X \in \mathbb{R}^{d\times K}$ is the matrix formed using these parameter vectors. 
The learning process is performed to find a minimizer of the objective function, $x_i = x^{*} = \arg\min_{x \in \mathbb{R}^d} f(x)$.

The data obtained from CAVs are typically non-independent and non-identically distributed (non-IID). FedAvg faces challenges in realistic heterogeneous data settings, as a single global model may not perform well for individual vehicles, and multiple local updates can cause the updates to deviate from the global objective~\cite{collins2022fedavg}. Several variants of FedAvg have been proposed to address the challenges encountered by FL, such as data heterogeneity, client drift, local vehicle data imbalance, communication latency, and computation capabilities. FedProx algorithm, FedAvg with a proximal term, has been proposed to improve the convergence and reduce communication cost ~\cite{li_federated_2020}. Dynamic Federated Proximal~\cite{zeng_federated_2022} algorithm (DFP) is an extension of FedProx that could effectively deal with non-IID data distribution by dynamically varying the learning rate and regularization coefficient during the learning process. FedAdam~\cite{reddi_adaptive_2022} has shown improved convergence and optimization performance by incorporating ADAM optimization in the FedAvg algorithm. \textcolor{black}{Improving the performance of the FL model is an ongoing research activity~\cite{acar_federated_2022,li_convergence_2020,wang_tackling_nodate,wang_field_2021}.}

\subsubsection{Knowledge Distillation}
In this subsection, we discuss the integration of knowledge distillation with FL. Federated Distillation (FD)~\cite{jeong_communication-efficient_2018} uses knowledge distillation to transfer knowledge in a decentralized manner, leading to a significant reduction in the communication size compared to a traditional FL. It also has the ability to handle non-IID data samples~\cite{liu_communication-efficient_2022}. Wang~\etal ~\cite{wang_federated_2022}  proposed a conceptual framework called FD for CAV (FDCAV), where CAVs share their outputs (e.g., bounding boxes) with a central server, which computes the average output from the global model and sends it back to vehicles. The vehicles then update their local models based on the output of the global model~\cite{wang_federated_2022}.

Another approach is to deploy a teacher model on the server and student models on the clients. In this process, client devices usually train and deploy a smaller, simpler model to mimic the behavior of a larger, more complex model residing on the server. It allows for the transfer of knowledge from the larger server model to the smaller client model, thereby reducing computational complexity and enhancing efficiency. For example, in Federated Group Knowledge Transfer (FedGKT)~\cite{he2020group}, a ResNet-55 or ResNet-109 is deployed on the server, while a ResNet-8 is utilized on the clients. Similarly, Federated Knowledge Distillation (FedKD)~\cite{wu2022communication} employs a comparable approach, conducting experiments on natural language recognition tasks. Knowledge distillation with FL is particularly beneficial in scenarios where computational resources or storage capacities are constrained or where the deployment of larger models is infeasible. CAVs are prime examples of such application scenarios. \textcolor{black}{}{The CFL is summarized in Algorithm~\ref{alg:CFL}. }

\begin{algorithm}[t]
\footnotesize
    \caption{CFL for Dynamic Data Updating CAV}
    \label{alg:CFL}
    \begin{algorithmic}[1]
    \renewcommand{\algorithmicrequire}{\textbf{Input:}}
    \renewcommand{\algorithmicensure}{\textbf{Output:}}
        \REQUIRE Vehicle set $\mathbb{V}$, communication rounds $T$, isolated time-varying local dataset $\xi = \{\xi_v^{(t)} : v \in \mathbb{V}\}$, local epochs $E$, learning rate $\{ \eta_t \}_{t=0}^{T-1}$, loss function $f$
        \ENSURE Aggregated global model $\theta$
        \STATE For each vehicle $v \in \mathbb{V}$ initialize model: $\theta_v^{(0)} \  \in \ \mathbb{R}^d$
        \FOR{$t = 0, \dots, T-1$} 
        \STATE {\bfseries Perform} local SGD {\bfseries for} vehicle $v \in \mathbb{V} $ {\bfseries in parallel do}
            \STATE \ \ \ Sample $\xi_v^{(t)} \mbox{, compute } g_v^{(t)} := \widetilde{\nabla} f_v (\theta_v^{(t)},\xi_v^{(t)})$
            \STATE \ \ \ $ \theta_v^{(t+1)} \ \gets \  \theta_v^{(t)} - \eta_t g_v^{(t)} \hspace{0.5cm} \implies $ SGD ($E$ epochs)
            \STATE \ \ \ Vehicle sent model $\theta_v$ to server
        \STATE {\bfseries end for}
        \STATE $ \theta^{(t+1)} \ \gets \ \sum_{v \in \mathbb{V} } \frac{|\xi_v^{(t)}|}{|\xi^{(t)}|} ( \theta_v^{t+1} ) \hspace{0.12cm} \implies $ Aggregation on server
        \STATE Server sent model $\theta^{(t+1)}$ to vehicles
       \ENDFOR
    \STATE Output the aggregated global model $\theta \gets \theta^{(T)}$
    \end{algorithmic}
\end{algorithm}

\subsection{Decentralized Federated Learning} 
\label{Sec. DFL for CAV}

In the CFL paradigm, model parameters (weights or gradients) are transmitted to a central server, often a Road-Side Unit (RSU), where the FL server-side aggregation process takes place. On the contrary, DFL relies on a consensus among the vehicles, fostering collaboration to collectively update global parameters without the need for a central server. \textcolor{black}{The DFL algorithm is shown in Algorithm~\ref{alg:DFL2}.}
The scalability of CFL is limited by the computational capacity of the server, which requires a dedicated infrastructure. The dependence on a single server introduces a potential point of failure in the learning process and can lead to communication congestion between the server and vehicles, especially when handling a substantial number of vehicles~\cite{nguyen2022deep}. 

DFL offers scalability by accommodating a large number of vehicle clients without relying on a central server, 
and exhibits enhanced robustness since the collaborative training among vehicles can continue even if an individual vehicle becomes unavailable. DFL relies on the V2X communication module to send model data directly to other neighboring vehicles for updates~\cite{lu_collaborative_nodate,pokhrel_decentralized_2020}.  

\begin{figure}[b]
    \centering
    \includegraphics[width=8.5cm]{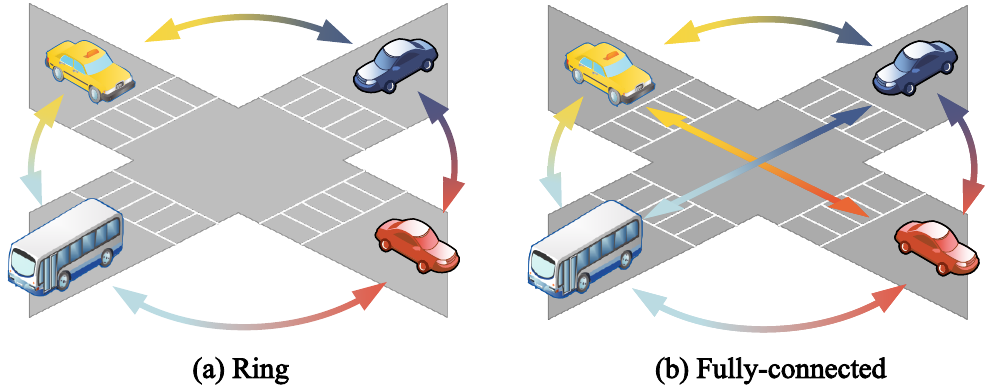}
    \caption{Ring (\textit{left}) and fully-connected topology (\textit{right})---four vehicles.}
    \label{fig:topology}
\end{figure}

The primary concept behind the DFL process is to establish consensus among vehicles by enabling communication exclusively between adjacent neighbors. This communication process can be effectively represented by employing a consensus/gossip matrix within a network topology graph. \textcolor{black}{More precisely, a vehicle $i$ communicates with vehicle $j$ based on a non-negative weight that represents the connectivity of vehicle $i$ and vehicle $j$, that is, $w_{ij} > 0$.} The case $w_{ij} = 0$ indicates that no communication takes place between $i$ and $j$. Similarly, for self-loops, the associated weight is represented by $w_{ii} > 0$.
Fig.~\ref{fig:topology} shows examples of two commonly employed network topologies, namely the ring and the fully connected for the $n=4$ client/vehicle configuration. In the fully connected topology setting, all vehicles interact with each other, whereas in the ring topology, vehicles interact only with neighboring vehicles. These associated weights can be compiled into a matrix of dimension $n \times n$ and can be written as $W = [w_{ij}] \in [0,1]^{n\times n}$. The most standard name for $W$ used in the literature is \textit{gossip} or \textit{mixing matrix}.

\begin{algorithm}[t]
\footnotesize
    \caption{DFL for Dynamic Data Updating CAV}
    \label{alg:DFL2}
    \begin{algorithmic}[1]
    \renewcommand{\algorithmicrequire}{\textbf{Input:}}
    \renewcommand{\algorithmicensure}{\textbf{Output:}}
        \REQUIRE Vehicle set $\mathbb{V}$, communication rounds $T$, isolated time-varying local dataset $\xi = \{\xi_v^{(t)} : v \in \mathbb{V}\}$, local epochs $E$, learning rate $\{ \eta_t \}_{t=0}^{T-1}$, loss function $f$, mixing matrix $W$
        \ENSURE Personalized model $\theta_v$ for each vehicle $v \in \mathbb{V}$
       
        \STATE For each vehicle $v \in \mathbb{V}$ initialize model: $\theta_v^{(0)} \  \in \ \mathbb{R}^d$
       \FOR{$t = 0, \dots, T-1$}
        \STATE {\bfseries Perform} local SGD {\bfseries for} vehicle $v \in \mathbb{V} $ {\bfseries in parallel do}
            \STATE \ \ \ Sample $\xi_{v}^{(t)} \mbox{, compute } g_v^{(t)} \gets \widetilde{\nabla} f_v (\theta_v^{(t)},\xi_v^{(t)})$
            \STATE \ \ \ $ \theta_v^{(t+\frac{1}{2})} \ \gets \  \theta_v^{(t)} - \eta_t g_v^{(t)}   \implies $ SGD ($E$ epochs) 
            \STATE Vehicle sent model $\theta_v^{(t+\frac{1}{2})}$ to other vehicles
        \STATE {\bfseries end for}  
        \STATE  Aggregate models of other vehicles $ u \in \mathbb{V}$: \\
            $ \theta_v^{(t+1)} \ \gets \ \sum_u W \ \theta_u^{(t+\frac{1}{2})} \implies $ Aggregation on clients
        \ENDFOR
        \STATE Each client deploys a personalized model $\theta_v \gets \theta_v^{(T)}$
    \end{algorithmic}
\end{algorithm}

The mixing matrix, $W = [w_{ij}] \in [0,1]^{n\times n}$, is a non-negative, symmetric $(W = W^{\top})$ and doubly stochastic, that is, $W\mathds{1} = \mathds{1}, \mathds{1}^{\top}W = \mathds{1}^{\top}$ matrix, where $\mathds{1}$ is the column of ones. Then, the consensus operation can be represented as,
\\
\begin{equation}
    \label{eq:consensus-step}
    \theta_i^{(t+1)} \ = \ \sum_{j \in [n] } w_{ij}^{(t)} \  \theta_j^{(t)} ,
\end{equation}
where $\theta$ is the model parameter (weights/gradients). 

\textcolor{black}{However, DFL also encounters notable limitations, including hindered convergence (caused by the heterogeneity of data) and network latency, and the need to synchronize/arbitrate parameters and adapt to dynamic network topologies during vehicle communications}. These challenges arise from the decentralized nature of the FL framework, which requires efficient mechanisms to address disparities in data distribution and network connectivity among the participating vehicles~
\cite{li2022gcn,barbieri2022decentralized,liu2022enhanced,beltran2022decentralized,wilbur2020time}.

\section{Overview of Data Modalities, Base Machine Learning Models, and Securities}
\label{Sec. Overview}

The concept of FL4CAV is illustrated in Fig.~\ref{Fig. FL_CAV}. Each CAV as a client, undertakes sensing data acquisition, signal processing, storage, communication, perception, and decision-making. For sensing data acquisition, a variety of sensors are integrated into CAVs, including Global Navigation Satellite Systems (GNSS), multi-modal cameras, Radio Detection And Ranging (Radar), Light Detection And Ranging (LiDAR), and Inertial Measurement Unit (IMU) to capture the vehicle, driver, passenger, and external information. 

\begin{figure*}[t]
\centering
\includegraphics[width=1\linewidth]{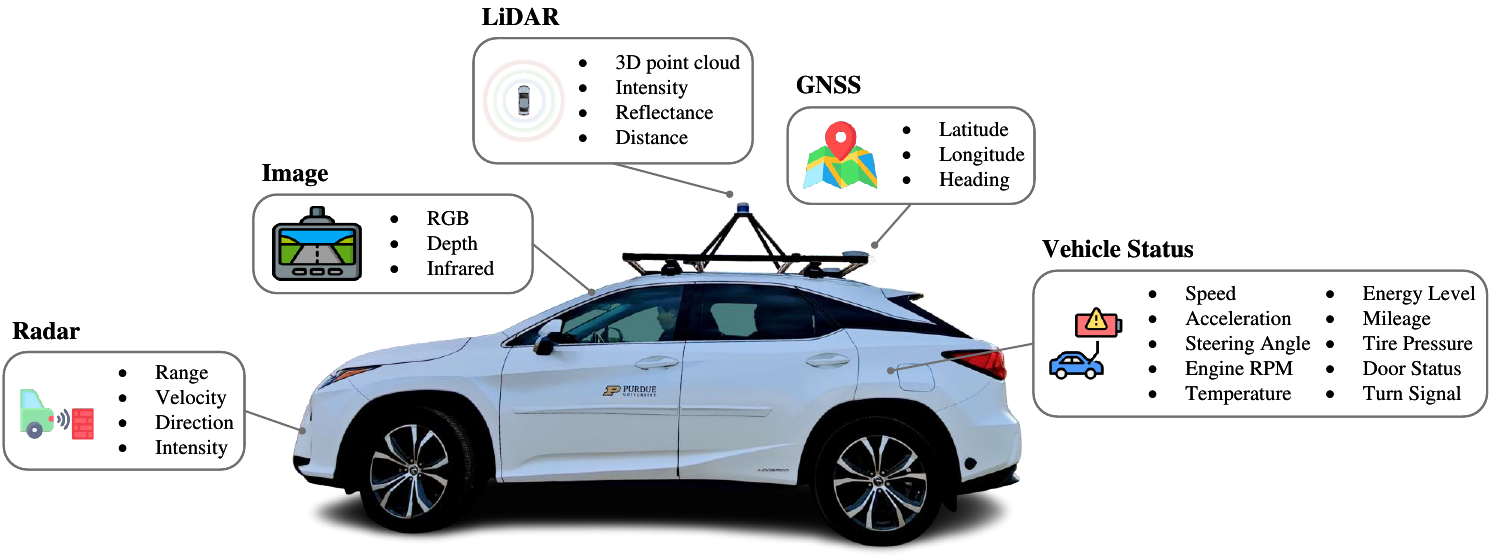}
\caption{Illustration of various data sources from a connected and automated vehicle.}
\label{Fig. Data}
\end{figure*}

CAV tasks are diversified to include tracking the target speed, prediction of behavior, motion planning, motion control, object detection, and in-vehicle human monitoring. After training on ML models with local data, clients send the trained model to the server. Then, the server shares a generalized model with clients for perception, prediction, and decision-making purposes. The FL4CAV framework shows a trend towards multi-modal sensing data, massively parallel clients, and multi-class tasks.

An overview of the data modalities, the base ML models of CAVs, and data security is presented next.

\subsection{Data Modality}

\textcolor{black}{CAVs collect multi-modal data from various sensors to perform tasks such as navigation and perception. The impact of different types of data modalities during the FL process on sensor fusion is dynamically diverse~\cite{yuan2023fedmfs}}. The data collected by sensors depend on the sensor type, the sensor's range, the accuracy/precision of the sensor, sensor placement, and the operating environment. The operating environment, such as snow, heavy rain, or fog, can reduce sensor visibility, thereby deteriorating data quality. These factors lead to variations that can significantly affect the sensor performance. The performance of the FL model is directly dependent on the quality of the data collected by the vehicles. The data resolution, size, and sampling rate obtained from CAVs are generally heterogeneous, and processing the data is also a challenging task. In the following, we review the various data modalities in FL4CAV applications that are illustrated in Fig.~\ref{Fig. Data}.

\subsubsection{Image} Images, especially visible RGB images, are one of the most important data modalities for CAVs. Vision-related tasks, such as driver monitoring~\ref{Sec. Driver Monitoring}, steering wheel angle prediction~\ref{sec-Steering_wheel_appl}, object detection~\ref{Sec-obj_detection}, traffic sign recognition~\cite{stergiou_federated_2022}, and semantic segmentation~\cite{fantauzzo_feddrive_2022} use images captured by the camera as the data source. In most applications, various ML models are trained to achieve the intended functionality. However, due to its intrusive design, privacy issues are always a concern for image-based systems, especially for in-cabin and driver-related applications \cite{yuan2021smart,mishra_cabin_2022,doshi2022federated,yuan2023federated,zhao2023fedsup}. Privacy concerns for visual image-based systems are addressed by FL since only the model parameters are transmitted while the user data are kept locally in the vehicle. Moreover, FL also solves the data transmission problem due to the large size of images and video data, thus leading to a more communication-efficient learning framework.

\subsubsection{LiDAR} \textcolor{black}{LiDAR data is vital for automated driving capabilities, which has been used for object detection tasks~\cite{elbir2022federated,liu2022parallel2} and cooperative perception scenarios~\cite{xu2023v2v4real, ma2023macp}. LiDAR generates 3D point clouds that can detect objects accurately even under adverse weather conditions, unlike camera data that are unreliable under similar conditions}. However, the dense point cloud of LiDAR data makes transmission a demanding task. FL system for LiDAR data can improve learning efficiency and save communication resources while being able to handle large data sets.

\subsubsection{Radar} 
Radar sensors are used for object detection and collision avoidance in applications such as automatic emergency braking, traffic alerts, and adaptive cruise control~\cite{pandey2022classification,venon2022millimeter,liu2022parallel,cui2023red}. Radars have long operating ranges, good measurement accuracy, and are operational in varying weather conditions~\cite{bilik2019rise}. Radar data provides critical information about the vehicle's surroundings, including the position and the speed of other objects. Similarly to LiDAR, the FL system for Radar can also improve learning efficiency and save communication resources.

\subsubsection{Vehicle Status and GNSS} Vehicle status data such as velocity, acceleration, throttle/brake command, vehicle global position through GNSS, and other vehicle parameters are also an important part of the CAV data modality. These parameters are relevant primarily to the vehicle rather than to the external environment. These data typically reveal sensitive information about driver locations, habits, and behaviors that could potentially compromise their privacy and security. FL addresses these privacy concerns well while utilizing these data to improve several applications such as collision avoidance~\cite{fu_selective_2022}, vehicle trajectory prediction~\ref{sec-veh_traj_appl}, and motion control application~\ref{sec-motion_control_appl}.

\subsection{Base Models in FL4CAV Applications}
\label{Sec.DeepLearning}

ML has been widely used to achieve superior performance in various complex tasks, given the availability of multi-modal data from in-vehicle sensors. Furthermore, the ML in FL4CAV shows the feasibility of implementation in real time, which is required due to the limited computing and communication resources of vehicle equipment. We next discuss the various ML architectures that are used as base models in critical tasks of CAVs.  

\subsubsection{Multilayer Perceptron} A Multilayer Perceptron (MLP), as a classic ML architecture, consists of multiple layers of fully connected neurons. It can be applied to various vehicle-related tasks, including perception, decision-making, and control. MLP provides a flexible and versatile tool for modeling complex relationships in vehicle-related data. However, its performance in specific tasks may be limited due to the computational demand for large models. \textcolor{black}{Due to their applicability, MLPs are widely employed in vehicle trajectory prediction (Sec. \ref{sec-veh_traj_appl}), motion control (Sec. \ref{sec-motion_control_appl}), and traffic flow prediction (Sec. \ref{Sec-traffic-prediction}).}

\subsubsection{Convolutional Neural Network}
Convolutional Neural Networks (CNNs) are presently one of the most popular architectures in ML. They are known for their excellent performance in handling image-related tasks. CNN uses convolutional layers to automatically extract features from images and learn to associate these features with corresponding labels. \textcolor{black}{CNNs exhibit versatile performance in performing a wide range of tasks, including, but not limited to, classification (as exemplified by LeNet~\cite{lecun1998gradient}, ResNet~\cite{he2016deep}), object detection (such as the YOLO~\cite{redmon2016you} framework), and mask generation for semantic segmentation (represented by models such as U-Net~\cite{ronneberger2015u}, BiSeNet~\cite{yu2018bisenet}), among others.} Due to its high efficiency in extracting features from image data, CNNs are widely applied in various vehicle-related applications, such as in-vehicle human monitoring (Sec. \ref{Sec. Driver Monitoring}), steering wheel angle prediction (Sec. \ref{sec-Steering_wheel_appl}),  object recognition (Sec. \ref{Sec-obj_detection}).

\subsubsection{Recurrent Neural Network}
Recurrent Neural Networks (RNNs) excel at extracting spatial relationships in features. They are specifically designed to capture temporal dependencies in sequences of data. Some popular RNN architectures include Long Short-Term Memory (LSTM)~\cite{hochreiter1997long} and Gated Recurrent Unit (GRU)~\cite{cho2014learning}. In the context of vehicles, RNNs have found extensive applications in modeling the motion and behavior of vehicles, their surroundings, and targets. \textcolor{black}{Using their sensitivity to time series data, RNNs can effectively capture the dynamics and temporal patterns in various vehicle-related scenarios, for example vehicle trajectory prediction (Sect. \ref{sec-veh_traj_appl}) and traffic flow prediction (Sec. \ref{Sec-traffic-prediction}).}

\subsubsection{Transformer}
Transformer~\cite{vaswani2017attention} architecture and its variant, Vision Transformer (ViT)~\cite{dosovitskiy2020image}, have emerged as powerful alternatives to traditional CNNs and RNNs. The Transformer architecture, initially introduced for natural language processing tasks, has shown exceptional performance in various domains, including computer vision. Transformers take advantage of self-attention mechanisms to capture global dependencies across the input sequence or image. This allows them to effectively model long-range dependencies and contextual relationships, leading to improved performance in tasks such as image classification, object detection, and semantic segmentation. Transformers' ability to capture global context and long-range dependencies makes them well-suited for various tasks in the automotive domain~\cite{tian2022federated}. \textcolor{black}{Transformers and ViTs have attracted substantial attention in the fields of FL \cite{li2023fedtp} and CAV \cite{xu2022v2x} due to their ability to effectively capture global information. Transformers and ViTs have a potential for a wide range of vehicular applications (Sec. \ref{Sec. Applications of FL for CAV}).}

\subsubsection{Generative Network} Generative networks form images based on input data, such as mask labels and super-resolution. \textcolor{black}{These networks, also known as Generative Adversarial Networks (GANs)~\cite{goodfellow2014generative} or Variational Auto-Encoders (VAEs)~\cite{kingma2013auto}, exhibit a distinctive ability to generate high-quality and realistic images. Generative networks have attracted attention in both the FL~\cite{zhang2023novel} and CAV~\cite{roy2019vehicle} domains}. \textcolor{black}{However, there is still no unified framework that incorporates all three technologies. With the extensive application of generative networks in CAV, combined with FL's enhancement of privacy protection and learning efficiency, they have a potential for various applications. In vehicular applications, generative networks provide several use cases, such as vehicle trajectory prediction (Sec.~\ref{sec-veh_traj_appl}). One application lies in super-resolution, where generative networks can enhance the resolution and details of low-resolution images, proving useful for tasks such as license plate recognition and surveillance systems. }Furthermore, generative networks can also be utilized to augment and improve data sets in training data sets for vehicle-related tasks.

\subsubsection{Reinforcement Learning}
\textcolor{black}{Reinforcement learning (RL) demonstrated superior capabilities in solving complex decision-making problems, surpassing human-level performance in various domains~\cite{sutton2018reinforcement}}. RL improves the abilities of the agent through interaction with the environment, enabling the agent to learn optimal policies through trial and error. RL has been extensively applied in CAV operations, such as motion control (Sec. \ref{sec-motion_control_appl}), vehicle trajectory prediction (Sec. \ref{sec-veh_traj_appl}),vehicular CPS (Sec. \ref{Sec-veh_network}), and resource allocation~\cite{li2022lane,selvaraj2022ml,lu2023event}.

\subsection{Model Security}
 
Robust and secure privacy preservation techniques are essential to protect sensitive data during the FL training process for CAVs. It is demonstrated that the training can still be vulnerable to various malicious attacks, such as when one or more participants are compromised, and they could transmit false parameters to hinder the global model performance. The FL central server is also prone to attacks that may cause the entire learning process to collapse~\cite{al2021cybersecurity}. The type of data considered in this section refers to the model parameters, such as gradients or weights, that are transmitted to the server/neighboring vehicles. These are not the raw data used for the training of the local model, as they are inherently preserved in the FL process.

Homomorphic encryption, differential privacy, and blockchain-based techniques are notable methods to preserve privacy in FL4CAV. These approaches aim to minimize the trade-offs between model performance and data privacy, ensuring data security while enabling effective model performance. A review of various cyber-security threats can be found in \cite{ju2022survey,hussain2022cyber,ghimire2022recent,alazab2021federated,ferrag2021federated,pandey2022review}. We will next discuss some of the widely used privacy-preserving techniques. 

\subsubsection{Homomorphic encryption}

Homomorphic Encryption (HE) is a powerful technique that allows the server to perform training on encrypted vehicle data without the need for decryption, thus ensuring data privacy and security. In particular, it allows direct computation on encrypted data with decrypted results~\cite{peng2021bflp}. 

\subsubsection{Differential Privacy}

Differential Privacy (DP) is an approach that safeguards data privacy by injecting random noise into the data before transmitting them to the server, preventing unauthorized extraction of sensitive information while also preserving data ownership and alignment with regulatory compliance. However, there is a trade-off between privacy settings and accuracy that can impact the performance of the models. DP has been used in multiple applications of FL4CAV for incorporating data security~\cite{chen_bdfl_2021,basnet2021deep, korba2023federated,hussaincyber,sani2022privacy}

\subsubsection{Secure Multi-Party Computation}

Secure Multi-Party Computation (SMPC) employs cryptographic techniques to encrypt and partition the data, enabling collaborative computation on these data with output results accessible to vehicles. While SMPC introduces considerable communication overheads, it excels in preserving data security and privacy. The SMPC integration framework with FL facilitates model training on data without exposing model parameters~\cite{byrd2020differentially,bonawitz2016practical}. In an FL setup, SMPC can substantially encrypt the communication of model updates. Furthermore, vehicles can apply SMPC for collaborative intermediate computations, ensuring that even if the central server is compromised, no meaningful information can be derived from these computations.

\subsubsection{Physical Security}

\textcolor{black}{In model security enhancement, reinforcement at the physical layer emerges is a key approach. Beyond the security measures required for both the vehicle and the server, hardware-level security technologies, such as the Trusted Execution Environment (TEE), offer participants in FL an isolated, secure, and confidential execution environment~\cite{mo2021ppfl}. With the support of TEE, vehicles can process and store data within a protected environment, iterating and refining their local models. Similarly, the server, using the support of TEE, can execute model aggregations in a secure context.}

\subsubsection{Blockchain}

Another disruptive technology gaining traction in CAV applications is blockchain-based methods, leveraging the decentralized and tamper-resistant nature of blockchain to improve data integrity, transparency, and security~\cite{singh_blockchain_2017,rathee_blockchain_2019,fu_vehicular_2020,pokhrel_federated_2020,basha_inter-locking_2021,he_bift_2022,javed_integration_2022,zhu2022crowdsensing}. Blockchain is a type of digital ledger technology that securely transfers data in a decentralized framework. CAVs share their data with the vehicular network and the information is stored on the blockchain. Blockchain provides a secure, credible, and decentralized approach to FL, enabling collaborative model training while safeguarding data privacy \cite{zhang2023cfsl}. The system is designed to protect data privacy and security, as well as to provide greater security to the general vehicular networks involved in the learning process~\cite{dorri_blockchain_2017}. An analysis of various privacy preservation approaches is given in~\cite{hussain2022cyber,billah2022systematic}.

In FL4CAV, the model parameters of individual vehicles can be stored as transactions on the blockchain, ensuring transparency and accountability. This creates trust among the vehicles, as the model updates can be verified. Additionally, blockchain enables incentive mechanisms through smart contracts, which reward CAVs that contribute high-quality model updates or share their computational resources for training. These incentives encourage active participation and foster collaboration among vehicles~\cite{lu2020blockchain,peng2021bflp,liu2021blockchain,he2021blockchain}.

\begin{table*}[!t]
\caption{Literature Overview of FL for CAV Algorithms}
\label{CFL Table Related literature}
\centering
\begin{tabularx}{\linewidth}{|l|l|l|l|p{70pt}|X|}
\hline
\textbf{Literature} & \textbf{Time} & \textbf{CFL} & \textbf{DFL} & \textbf{Base Model} & \textbf{FL Algorithm} \\

\hline
Doomra \etal \cite{doomra_turn_2020} & 2020 & \cmark & & LSTM & Averaging \\
\hline
Liu \etal \cite{liu2020privacy} & 2020 & \cmark & & GRU & Averaging (randomly client selection aggregation.) \\
\hline
Zhang \etal \cite{zhang_end--end_2021} & 2021 & \cmark & & Two-stream CNN & Averaging \\
\hline
Aparna \etal \cite{m_p_steering_2021} & 2021 & \cmark & & CNN & Averaging\\
\hline
Rjoub \etal \cite{rjoub_improving_2021} & 2021 & \cmark & & YOLO & Averaging\\
\hline
Kong \etal \cite{kong2021FedVCP} & 2021 & \cmark & & MLP & Averaging \\
\hline
Zhou \etal \cite{zhou2021two} & 2021 & \cmark & & CNN & Averaging (hierarchical) \\
\hline
Saputra \etal \cite{saputra2021dynamic} & 2021 & \cmark & & MLP & Averaging (optimal economic client selection aggregation) \\
\hline
Barbieri \etal \cite{barbieri2021decentralized} & 2021 &  & \cmark & PointNet & Mesh topology \\
\hline
Zeng \etal \cite{zeng_federated_2022} & 2022 & \cmark & & MLP & Averaging \\
\hline
Stergiou \etal \cite{stergiou_federated_2022} & 2022 & \cmark & & LeNet-5 & Averaging \\
\hline
Fantauzzo \etal \cite{fantauzzo_feddrive_2022} & 2022 & \cmark & & BiSeNet V2 & Averaging \\
\hline
Elbir\etal \cite{elbir2022hybrid} & 2022 & \cmark & & U-Net & Averaging (hybrid federated and centralized learning architecture) \\
\hline
Han \etal \cite{han_federated_2022} & 2022 & \cmark & & LSTM & Averaging \\
\hline
Fu \etal \cite{fu_selective_2022} & 2022 & \cmark & & RL & Averaging (reputation, quality, and overhead client selection aggregation) \\
\hline
Doshi \etal \cite{doshi2022federated} & 2022 & \cmark & & ResNet-8 \& 56 & Knowledge Distillation (FedGKT) \\

\hline
Sepasgozar \etal \cite{sepasgozar2022fed} & 2022 & \cmark & & LSTM & Averaging \\

\hline
Yang \etal \cite{yang2022efficient} & 2022 & \cmark & & ResNet-18 & Averaging (partial model weight update) \\
\hline
Zhou \etal \cite{zhou2022stfl} & 2022 & \cmark & & Transformer & Averaging (spatial and temporal client selection aggregation) \\
\hline
Yuan \etal \cite{yuan2023federated} & 2023 & \cmark & & ResNet-34 & Averaging (selective aggregation; meta-learning personalization) \\
\hline
Yuan \etal \cite{yuan2023peer} & 2023 & & \cmark & ResNet-34 & Gossip protocol \\
\hline
Zhao \etal \cite{zhao2023fedsup} & 2023 & \cmark & & CNN & Averaging (hierarchical) \\
\hline
Du \etal \cite{du2023driver} & 2023 & \cmark & & LSTM & Averaging (hierarchical) \\
\hline
Parekh \etal \cite{parekh2023gefl} & 2023 & \cmark & & CNN & Averaging \\
\hline
Wang \etal \cite{wang2023federated} & 2023 & \cmark & & LSTM & Averaging \\
\hline
\end{tabularx}
\end{table*}

\begin{table*}[!t]
\caption{Literature Overview of FL for CAV Applications}
\label{Table Related literature}
\centering
\begin{tabularx}{\linewidth}{|l|l|p{90pt}|p{90pt}|X|}
\hline
\textbf{Literature} & \textbf{Time} & \textbf{Data Modality} & \textbf{Application} & \textbf{Dataset} \\

\hline
Doomra \etal \cite{doomra_turn_2020} & 2020 &  Time series data of multiple features from sensors  & Turn signal prediction & Ford’s Big Data Drive \cite{doomra_turn_2020} \\
\hline
Liu \etal \cite{liu2020privacy} & 2020 & Traffic flow & Traffic flow prediction & Caltrans Performance Measurement System (PeMS) dataset \cite{chen2002freeway} \\
\hline
Zhang \etal \cite{zhang_end--end_2021} & 2021 & RGB image & Steering angle prediction & Self-collected \\
\hline
Aparna \etal \cite{m_p_steering_2021} & 2021 & RGB image & Steering angle prediction & Self-collected\\
\hline
Rjoub \etal \cite{rjoub_improving_2021} & 2021 & RGB image and LiDAR & Object detection & Canadian Adverse Driving Conditions Dataset \cite{pitropov_canadian_2021}\\
\hline
Kong \etal \cite{kong2021FedVCP} & 2021 & Trajectory data & Cooperative positioning & Didi Chuxing GAIA Initiative \cite{GAIA_Initiative} \\
\hline
Zhou \etal \cite{zhou2021two} & 2021 & RGB image & Traffic sign recognition & BelgiumTS \cite{timofte2014multi} \\
\hline
Saputra \etal \cite{saputra2021dynamic} & 2021 & Traffic accident data & Traffic accident prediction & 1.6 million UK traffic accidents \cite{UK_traffic_accidents} \\
\hline
Barbieri \etal \cite{barbieri2021decentralized} & 2021 & RGB image and Sensor data & Object detection & nuScenes dataset \cite{caesar2020nuscenes} \\
\hline
Zeng \etal \cite{zeng_federated_2022} & 2022 & RGB image and trajectory data  & Target speed tracking & Berkeley deep drive \cite{yu2020bdd100k} and dataset of annotated car trajectories \cite{moosavi2017trajectory} \\
\hline
Stergiou \etal \cite{stergiou_federated_2022} & 2022 & RGB image & Traffic sign recognition & German Traffic Sign Recognition Benchmark \cite{stallkamp2011german} \\
\hline
Fantauzzo \etal \cite{fantauzzo_feddrive_2022} & 2022 &  Multi-modal image  & Semantic segmentation & Cityscapes \cite{cordts_cityscapes_2016} and IDDA \cite{alberti_idda_2020-1}  \\
\hline
Elbir\etal \cite{elbir2022hybrid} & 2022 & RGB image and LiDAR & 3D object detection & Lyft Level 5 dataset \cite{kesten2019lyft} \\
\hline
Han \etal \cite{han_federated_2022} & 2022 & Vehicle Status & Trajectory prediction & US‐101 and I‐80 data sets of NGSIM \cite{dot2018next} \\
\hline
Fu \etal \cite{fu_selective_2022} & 2022 & Vehicle position, velocity and acceleration & Collision avoidance & Self-generated \\
\hline
Doshi \etal \cite{doshi2022federated} & 2022 & RGB image & Driver activity recognition & State Farm Distracted Driver Detection \cite{farm_2016} and AI City Challenge 2022 \cite{Naphade22AIC22} \\

\hline
Sepasgozar \etal \cite{sepasgozar2022fed} & 2022 & Vehicle velocity & Traffic flow prediction & CRAWDAD Vehicular dataset \cite{fujimotocrawdad} \\

\hline
Yang \etal \cite{yang2022efficient} & 2022 & RGB image & Driver activity recognition & State Farm Distracted Driver Detection \cite{farm_2016} and YawDD \cite{yang2020driver} \\
\hline
Zhou \etal \cite{zhou2022stfl} & 2022 & Trajectory data & Trajectory prediction & Didi Chuxing GAIA Initiative \cite{GAIA_Initiative} \\
\hline
Yuan \etal \cite{yuan2023federated} & 2023 & RGB image & Driver activity recognition & State Farm Distracted Driver Detection \cite{farm_2016} and Drive\&Act \cite{martin2019drive} \\
\hline
Yuan \etal \cite{yuan2023peer} & 2023 & RGB image & Driver activity recognition & State Farm Distracted Driver Detection \cite{farm_2016} and The 7th AI City Challenge \cite{Naphade23AIC23} \\
\hline
Zhao \etal \cite{zhao2023fedsup} & 2023 & RGB image & Driver fatigue detection & Blinking Video Database \cite{pan2007eyeblink} and Eyeblink8 \cite{drutarovsky2014eye} 
\\
\hline
Du \etal \cite{du2023driver} & 2023 & 3D head position & Lane-change prediction & Self-collected \\
\hline
Parekh \etal \cite{parekh2023gefl} & 2023 & RGB image & Traffic sign recognition & German Traffic Sign Recognition Benchmark \cite{stallkamp2011german} \\
\hline
Wang \etal \cite{wang2023federated} & 2023 & Vehicle pose & Trajectory prediction & VeReMi \cite{van2018veremi} \\
\hline
\end{tabularx}
\end{table*}

\section{Applications of FL for CAV} 
\label{Sec. Applications of FL for CAV}

\textcolor{black}{In this section, we review some applications of FL in CAV. The FL4CAV literature, including FL configuration, data modalities, underlying models, applications, FL algorithm, and datasets, can be found in Tables~\ref{CFL Table Related literature} and \ref{Table Related literature}. The strengths of FL, such as protecting privacy, improving learning efficiency, improving generalization ability, and reducing communication overhead, resulted in several FL4CAV applications.}

\subsection{In-Vehicle Human Monitoring} 
\label{Sec. Driver Monitoring}
In-vehicle human monitoring is a critical issue for CAV and ITS. The in-vehicle human monitoring serves not just the driver but also extends to the other passenger monitoring in vehicles~\cite{mu2023human}. Beyond the application in commercial taxis, human monitoring becomes particularly critical in large public transportation modes such as buses, subways, ferries, and more, where adequate human personnel for service may be lacking. \textcolor{black}{Consequently, computer-aided monitoring programs can effectively offer superior service quality and protect passenger safety by handling tasks such as passenger counting, passenger traffic, detecting elderly falls, and emergency situations such as fires.}

FL significantly enhances privacy protection, enriches and diversifies knowledge, and improves learning efficiency, which makes it crucial for the application of human monitoring in the vehicle in the deployment of CAVs. Given the sensitivity of personal privacy and the rarity of traffic accidents, FL serves as a valuable tool in these contexts. FL has the potential to enhance the security of user data onboard while enabling knowledge transfer and ensuring the generalizability of the model. However, in human-related applications where data are highly heterogeneous and personalized, it can be challenging to balance the generalization ability of the model with the need for personalization to specific users~\cite{ferrari2023deep}.

Driver monitoring applications, such as distraction detection, are critical safety features that monitor driver stability and alertness and warn distracted drivers to apply safety-critical actions~\cite{hu2022review,ansari2022human,lu2022shared,ma2023m2dar,du2023driver}. \textcolor{black}{The computational and communication efficiency issues in driver activity recognition are addressed in~\cite{doshi2022federated} and a novel framework (FedGKT) was proposed to reduce communication bandwidth and asynchronous training requirements.} Driver privacy may be a greater concern than steering wheel angle prediction and object recognition, leading to FL's ability to be more highlighted in terms of privacy protection. However, the driver monitoring application is a highly personalized application where the driver's behavior is strongly associated with personal habits, emotions, cultural background, and even the interpretation of instructions. This user heterogeneity poses a challenge for FL systems. For human-related applications, such as driver monitoring, personalized FL is the dominant solution~\cite{yuan2023federated}. A DFL framework was proposed in~\cite{yuan2023peer} that incorporates a gossip protocol for knowledge dissemination. This framework not only achieves personalized models without requiring any additional processing, but also incorporates a knowledge dissemination technique that significantly accelerates the training process.

Passenger monitoring applications are an emerging research area that involves detecting passengers' intents to board and leave and warning of dangerous behavior in public transportation~\cite{liu2021automatic}. However, this field has not yet received much attention due to the lack of available datasets and the difficulty of monitoring multiple users simultaneously. Nevertheless, the ability of FL to integrate knowledge about public transportation and the growing demand for passenger monitoring makes FL a promising application in this area.

\subsection{Steering Wheel Angle Prediction} 
\label{sec-Steering_wheel_appl}

The prediction of the angle of the steering wheel has become a crucial feature of self-driving. The performance of ADAS features, such as lane keep assist and lane departure warning, is based on the prediction of the steering angle~\cite{gidado_survey_2020, bojarski2016end}. The steering wheel angle prediction is used to estimate the steering wheel rotation angle based on the input of road images. The prediction of the steering wheel angle manages the lateral positioning of the vehicle, even under challenging circumstances, such as on unpaved and unmarked roads. The steering wheel angle prediction needs to adapt to different driving and environmental conditions, and thus requires continuous model updates for high accuracy. 

FL achieves the above objectives by enabling several vehicles to collaborate in learning from new data and updating the model in a relatively short time. FL offers the benefit of continuous and collaborative learning, low communication overhead, and data security that is needed to develop a robust prediction model. 

It was demonstrated that FL can collectively train the prediction model while, at the same time, significantly reducing communication costs. The study presented in~\cite{zhang_end--end_2021} demonstrated a significant improvement in edge model quality through the use of FL in CAV. Specifically, the study involved predicting steering wheel angles using two modalities of data: images and optical flow. In~\cite{m_p_steering_2021}, the performance of FL and centralized learning in steering angle prediction was assessed under different levels of noise and the results were comparable. Furthermore, this study considered the implications of communication load and disruptions, providing a comprehensive evaluation of the systems. This makes FL suitable for applications involving an increasing number of CAVs, specifically for tasks such as steering wheel angle prediction. 

\subsection{Vehicle Trajectory Prediction} 
\label{sec-veh_traj_appl}
An accurate vehicle trajectory prediction allows CAVs to perform proper motion planning, as well as anticipate potentially dangerous behaviors of other vehicles, such as sudden lane change, skidding, or hard braking, react proactively and prevent accidents~\cite{sighencea2021review,liu2021survey,huang2022survey,liao2023driver,teng2023motion}. This is a challenging task and would require substantial amounts of sensitive vehicle data to train a model for trajectory prediction. 

FL is a viable solution that provides a collaborative learning framework with multiple vehicles while keeping sensitive local data private and secure. FL models are trained on diverse data from various vehicles operating in different scenarios. This enhances the generalization of the model and enables vehicles to handle rare events such as traffic accidents, adverse weather, and risky behaviors. Additionally, the FL framework supports continuous learning and model updates, allowing quick adaptation to dynamic traffic, road conditions, and unfamiliar scenarios. 

Trajectory prediction models commonly rely on time series data that encompass vehicle/passenger position, velocity, and acceleration. These models leverage the strength of deep neural networks, mainly RNNs, and Transformers, that have proven effective in predicting trajectories for various entities, including vehicles and pedestrians, while also capturing their behavioral patterns~\cite{wang2019exploring}. FL framework has been shown to be effective in learning spatio-temporal features with the Transformer model~\cite{zhou2022stfl} (or the LSTM model~\cite{majcherczyk2021flow}) while also protecting user privacy.  FL coupled with One-Class Support Vector Machine (OC-SVM) has been used to detect anomalous trajectories at traffic intersections~\cite{koetsier2022detection}. The reported findings indicate that the federated approach improves both the overall accuracy of anomaly detection and the benefit of individual data owners. 
FL has been reported to perform similarly to centralized learning ~\cite{han_federated_2022, rjoub_explainable_2022,wang2022atpfl}. Centralized learning requires that all data from the private vehicle be transferred to the central server for training, whereas the data are kept locally in the vehicle in the case of FL. 

\subsection{Object Recognition} 
\label{Sec-obj_detection}
Object recognition is one of the main functions of the visual perception system of CAVs intended to detect and localize objects using sensor data such as LiDAR and high-resolution image/video. These data are large in size and sensitive from a privacy point of view. As a result, there are
limitations to deploying robust detection models in a traditional centralized learning approach due to privacy and communication overhead. 
These concerns can be mitigated by using an FL-based approach for CAVs. FL can effectively help CAVs detect various objects in different driving scenarios, road types, traffic conditions, and weather types. FL enables the CAV framework to learn efficiently with low communication overhead, which is particularly advantageous when the volume of data is much larger than the size of the ML model while also ensuring the privacy of the data.

FL has already been used in computer vision-related tasks, such as developing safety hazard warning solutions in smart city applications~\cite{liu_fedvision_2020}. \textcolor{black}{The accuracy of object detection models is generally poor under adverse weather conditions such as snow and rain}. FL frameworks have been shown to improve detection accuracy~\cite{zhao2019object} and perform better than the centralized and gossip-decentralized models~\cite{rjoub_improving_2021}. \textcolor{black}{Recently, studies have been carried out to improve the performance of FL on complex tasks such as object detection~\cite{wang2022edge}. }
In~\cite{wang_federated_2022}, it has been shown that with multistage resource allocation and appropriate vehicle selection, FL performance improved significantly compared to traditional centralized learning and baseline FL approaches. 
In~\cite{barbieri2021decentralized}, a decentralized FL method is used for object classification using LiDAR on CAVs. The parameters of the ML model (PointNet~\cite{qi2017pointnet}) are communicated through V2V networks. It has been experimentally confirmed that FL is highly effective compared to self-learning approaches. 

Another important application of FL is the recognition and detection of license plates. It is used in ITS for applications such as traffic safety and violations, traffic monitoring, illegal/overtime parking detection, and parking access authentication. ML techniques have been shown to be highly efficient in detecting objects and recognizing license plates~\cite{weihong2020research,puarungroj2018thai,zang2015vehicle,zherzdev2018lprnet}. However, due to the large size of the data from all vehicles, it is not feasible to train on a real-time edge device. FL techniques offer numerous advantages to license plate detection and recognition systems, namely: privacy protection, enabling collaborative learning, and reduced network bandwidth requirements. These benefits of using FL contribute to increased effectiveness and adaptability of such systems in real-world scenarios~\cite{kong2021federated,xie2023asynchronous}.

\subsection{Motion Control} 
\label{sec-motion_control_appl}
\textcolor{black}{The motion controller of the vehicle executes the desired trajectory by determining the optimal control of the acceleration pedal position (longitudinal acceleration motion), the steering of the vehicle (lateral motion) and the brake position (longitudinal deceleration motion)~\cite{xiao2022resource,paden2016survey,tian2022parallel}.}
FL enables CAVs to train and optimize controller parameters collaboratively. \textcolor{black}{Some potential benefits of using FL are enabling CAVs to adapt to unseen routes/traffic scenarios or operating conditions due to previous data from other CAVs, acceleration on the ramp, driving in congested conditions, or challenges associated with higher vehicle speed~\cite{manna2022control}. FL enables CAVs to adapt to different driving scenarios, including unfamiliar roads, cities, and countries. Furthermore, FL may allow CAVs to adjust driving styles based on different driving habits, climates, scenarios, and cultural norms.}

FL has been used to dynamically update the controller parameters, resulting in improved achievement of the target speed with enhanced driver comfort and safety~\cite{zeng_federated_2022}. Additionally, FL finds application in collaborative optimization of control parameters between multiple vehicles at traffic intersections, resulting in the avoidance of collisions and improved driving comfort~\cite{wu2021traffic,prashanth2021hybrid}. \textcolor{black}{In~\cite{liu2021autonomous}, FL is utilized to improve brake performance under different driving conditions and environments by accurately determining road friction coefficients. This approach ensures the privacy of the driver while optimizing the braking action.} In~\cite{zeng_federated_2022}, an FL framework is proposed to optimize the controller design for CAVs with variable vehicle participation in the FL training process.

\textcolor{black}{Reinforcement Learning (RL) approach has been widely applied for motion control in vehicles due to its ability to train in complex scenarios with dynamic environments.} RL enables CAVs to learn control policies for the required objectives with user feedback and sensor measurements~\cite{li2022lane,chen2021deep,chen2020autonomous,folkers2019controlling,wang2018reinforcement}. \textcolor{black}{There are open research problems in motion control of CAVs that could be addressed by FL such as platooning, lane change maneuvers, merging on-ramps, signalized, and unsignalized intersections.} A review of existing CAV control methods is provided in~\cite{wang2018review,wang_survey_2020,liu2023unified}, while applications of ML to CAV control are reported in \cite{grigorescu2020survey,kuutti_survey_2019,li2019reinforcement,sharma2019lateral,wadekar2021towards,chellapandi2023predictivedoc}.

\subsection{Traffic Flow Prediction} 
\label{Sec-traffic-prediction}
Traffic flow prediction is one of the critical components of an ITS for efficient traffic control, safety, and management. Accurate predictions using historical data to forecast future traffic conditions can lead to reduced traffic congestion, such as optimal route recommendation and variable road signal timing. \textcolor{black}{Predicting traffic flow can also allow timely notification to authorities of occurrences of events, such as accidents and congested road conditions.} ML techniques, such as CNNs and RNNs, have shown promising results in predicting traffic flow~\cite{razali2021gap,sun2020machine,miglani2019deep}. 

FL has been used to predict traffic flow with improved accuracy while ensuring privacy and scalability. Sources for model training include data from CAV, RSUs, and traffic sensors. The predictions could be in real-time or for future time intervals, and the model can be trained to predict traffic patterns and improve the accuracy of traffic flow predictions. FL allows CAVs to collaboratively learn from their data while addressing privacy concerns.

\textcolor{black}{In~\cite{liu2020privacy}, a Gated Recurrent Unit (GRU) network is trained using FL to predict traffic flow. Experimental evaluations of a real-world data set show that the FL-based approach can achieve predictions comparable to those of traditional centralized approaches.} In~\cite{yuan2022fedstn}, an FL-based Spatial-Temporal Networks (FedSTN) algorithm was proposed to predict traffic flow. The algorithm employs various methods like Recurrent Longterm Capture Network, Attentive Mechanism Federated Network, and Semantic Capture Network (SCN) to learn spatial-temporal and semantic information.  It is reported that the FedSTN algorithm outperforms in terms of higher prediction accuracy compared to existing baselines such as Auto-Regressive Integrated Moving Average (ARIMA), eXtreme Gradient Boosting (XGBoost), FedGRU, and ST-ResNet~\cite{ zhang2017deep}. In~\cite{yuan2022fedtse}, a Long-Short-Term Memory (LSTM) is trained in an FL framework for traffic flow prediction along with an RL that is used for resource optimization. \textcolor{black}{In~\cite{sepasgozar2022fed}, an FL framework employing LSTM algorithm has been trained on a real Vehicular Ad hoc NETwork (VANET) data set based on V2V and V2R communication for the prediction of network traffic. The above developments show the benefits of using FL for complex tasks such as traffic flow prediction.}

\subsection{Vehicular Cyber-Physical Systems} 
\label{Sec-veh_network}

Vehicular Cyber-Physical Systems (VCPS) encompass the integration of physical systems, cyber systems, and vehicular communication networks~\cite{rawat2017vehicular}. Physical systems comprise vehicles, roads, and telematics/edge devices, while cyber systems include data centers, central servers (i.e., cloud), and traffic management systems. Vehicular networks, namely Cellular Vehicle-to-Everything (C-V2X) and V2X communication networks, play a key role in facilitating information sharing to improve driving comfort, safety, and traffic management. VCPS utilizes various technologies to enhance the vehicular network and enable seamless and robust communication between vehicles and systems.

FL plays a critical role in VCPS by enhancing data privacy and addressing resource constraints. FL uses a collaborative and distributed learning framework that captures data heterogeneity while eliminating the need to transfer local data from vehicles. This enables VCPS to benefit from FL's ability to preserve data privacy and facilitate efficient learning without compromising resource limitations.

In~\cite{lu2020federated}, an FL framework is proposed to detect and mitigate data leakage in VCPS while enhancing data privacy. The proposed scheme achieves good accuracy, efficiency, and high security based on simulations of a real-world data set. In~\cite{lei2022oes}, an FL framework (OES-Fed) is proposed for outlier detection and noise filtering in vehicular networks. In~\cite{zheng2022data}, extreme value theory (EVT) and personalized FL are proposed to model anomalous events caused by the non-heterogeneous data distribution among vehicles in vehicular networks. In~\cite{li2021federated}, an efficient and secure FL framework is combined with the Deep Q-Network (DQN) to ensure an efficient and secure scheme to reduce the latency of vehicular data sharing in vehicular networks.

\textcolor{black}{FL has gained significant acceptance for enhancing the resilience and robustness of VCPS networks against adversarial attacks}. This is achieved through the integration of FL with techniques such as differential privacy~\cite{olowononi2021federated} and blockchain-based approaches~\cite{he2021blockchain,ayaz2021blockchain}. These combinations have shown promising results in improving the security and reliability of the VCPS network.

\subsection{Vehicle-to-Everything Communication} 
\label{Sec-V2X}

An efficient and robust V2X communication such as V2V and V2I is a crucial step towards achieving an ITS. V2X communication plays a pivotal role in improving traffic management and enhancing driving comfort. 
\textcolor{black}{As ITS development progresses further, we expect a substantial increase in data transmission due to a large number of vehicles. This surge in data poses challenges in terms of communication and energy consumption. Moreover, given the private and sensitive nature of the data, ensuring data security is essential.} Therefore, it is crucial to address these issues by adopting energy-efficient approaches and establishing low-latency transmission in V2X communication~\cite{zhou2017energy}.

FL offers a promising solution for learning parameters with minimal latency and data transmission due to its decentralized training framework. It ensures data security while enabling efficient client/server selection during the training process~\cite{konevcny2016federated,li2020secure,song2023v2x} and resource management~\cite{prathiba2021federated,li2022federated}. These approaches have demonstrated an effective reduction in communication overload, addressing a significant challenge in FL implementations.

\textcolor{black}{In~\cite{samarakoon2018federated,samarakoon2019distributed}, extreme value theory was used in conjunction with an FL framework to model anomalous events, specifically large queue lengths. Lyapunov optimization was also incorporated for power allocation, which contributed to improving system performance.} 

\begin{figure*}[t]
\centering
\includegraphics[width=0.8\linewidth]{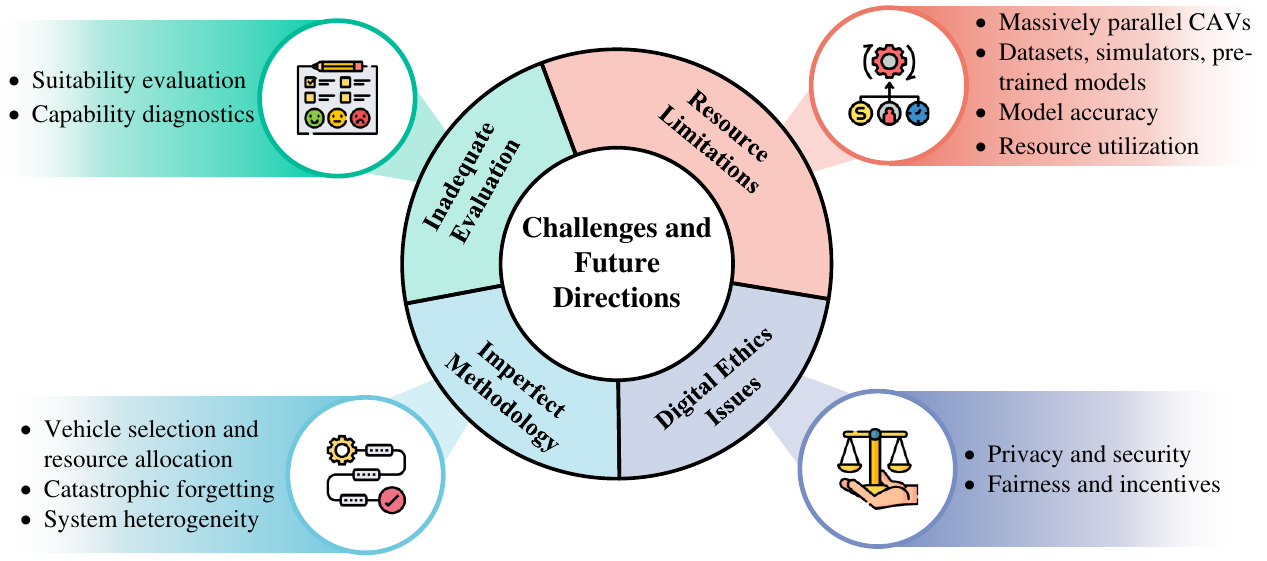}
\caption{Illustration of challenges and future directions of federated learning for connected and automated vehicles.}
\label{Fig. Challenge}
\end{figure*}

\section{Challenges and Future Directions} 
\label{Sec. Challenges}
In this section, we review various challenges in the use of FL4CAV as well as future research directions.

\subsection{Resource Limitations and Utilization}
\subsubsection{Collaboration capabilities and management in massively parallel CAVs} Significant participation of CAVs in FL could increase the solve time and memory utilization, and therefore calls for an increase in computational demand for a global model update. In particular, vision- and LiDAR-related perception tasks are characterized by large data sets that lead to high communication costs. Decentralized FL and clustered FL~\cite{taik_clustered_2022,hosseinalipour2023parallel,parasnis2023connectivity,fangsubmodel} are being explored to reduce communication overhead. 

The high communication demands and low reliability of 5G networks call for the development of 6G-V2X systems. Integrating 6G, V2X, and multi-access edge computing (MEC) powered by ML techniques creates the potential to achieve efficient and collaborative processing at the network edge. This approach aims to overcome the limitations of current 5G systems and pave the way for improved performance and reliability in future networks~\cite{yang2021edge,fang2021over}.

\subsubsection{Challenges due to lack of sufficient real-world datasets, simulators, and pre-trained base models} There is a need for more real-world datasets (different weather conditions and traffic scenarios), realistic high-fidelity FL4CAV simulators for seamless FL integration~\cite{wang_federated_2022,lobato2022flexe,dai2023online,hu2023cacc}, and good pre-trained models. 

\subsubsection{Low model accuracy}
FL often struggles with a trade-off between the accuracy achieved through model personalization and imposing high computational requirements on edge devices during learning. Split learning is one potential solution that enables efficient inference in resource-constrained edge clients while capturing both generalization and personalization capabilities~\cite{han2022splitgp}.

\subsubsection{Inefficient resource utilization}
Some of the issues of FL related to resource optimization include idle of powerful edge devices, underutilized network infrastructure, neglected edge devices without proper network connectivity, and discouraged sharing of parameters from edge devices with diverse privacy requirements~\cite{wang2023fair}. Therefore, there is a need for a robust FL framework that jointly utilizes and optimizes the resources of the device, server, and network infrastructure. 

Cooperative FL is a promising solution that overcomes these shortcomings and has been shown to be feasible and beneficial for learning processes leading to improved ML performance and resource efficiency~\cite{wang2023towards}. 
In another related study~\cite{zhang2022federated}, a cooperative architecture and an FL combined with an RL-based algorithm are proposed for the allocation of resources in CAV networks.

\subsection{Digital Ethics Issues}
\subsubsection{Privacy and security issues} Massive data also leads to privacy and security concerns. This problem must be addressed to train the ML model efficiently without compromising the model's accuracy and redundancy.

\subsubsection{Fairness and incentives} There is a need for appropriate rewarding policies and incentive mechanisms for CAVs to share the quality data needed for efficient model training performance~\cite{yuan2023digital}. 

\subsection{Imperfect Methodology}
\subsubsection{Lack of methods for efficient vehicle selection and resource allocation} Currently, there are no efficient methods that can filter useful data from CAVs to minimize network loading. There are ongoing efforts to develop reliable methods to optimally select vehicles and resource allocation schemes for efficient model training and communication ~\cite{wang_content-based_2021,tianqing2021resource,albelaihi_green_2022,fang2022communication}. In~\cite{nishio_client_2019}, the overall training process was demonstrated to be efficient when incorporating a client selection model. The setup looks at the resource availability of the clients and then determines the clients eligible to be part of the FL global model learning process.  In~\cite{rjoub_explainable_2022}, it is demonstrated that the model performance was improved with CAVs that were selected by trust-based deep RL.  

\subsubsection{Catastrophic forgetting} \textcolor{black}{CAVs cannot store all user data due to storage capacity limitations, and new data will always be generated during training iteration. }Therefore, when the FL framework is updated on new data in iteration, the global model might forget the previous knowledge, which may lead to catastrophic forgetting. This is another open research problem in FL4CAV.

\subsubsection{System heterogeneity in FL4CAV} Poor performance of the FL model (longer training time and a larger number of communication rounds) is generally caused by poor connectivity and slower devices (straggler devices). In traditional FL, a communication round is not complete until the data from all the chosen devices are available. \textcolor{black}{Hence, various adaptive strategies have been proposed to minimize the impact of stragglers and also eliminate them, if possible~\cite{wang_adaptive_2019,liu_adaptive_2023}.}

\subsection{Inadequate Evaluation Criteria}
\subsubsection{FL suitability evaluation for new users} It is often difficult for the newcomer vehicle to make any informed decisions. \textcolor{black}{In~\cite{rjoub_explainable_2022}, a trust-aware Deep RL model is proposed to assist new vehicles in making better trajectory and motion planning decisions. }
  
\subsubsection{Need for high capability diagnostics} There are several noise factors that could influence the decision of the FL, such as faulty sensors in a visual perception case and incorrect imputation of missing data. The development of robust diagnostics that can identify and eliminate the updates from these vehicles is needed. 

\section{Conclusions}
\label{Sec. Conclusion}
This survey paper reviews FL algorithms, data modalities, model security, and provides a list of critical applications and challenges of FL4CAV.
Currently, FL4CAV also presents unique challenges, such as ensuring data integrity, addressing communication latency, managing heterogeneous data sources, and maintaining model synchronization across different vehicles. However, with proper design and implementation, FL can offer significant advantages in terms of privacy preservation, network efficiency, and collaborative intelligence for CAVs.

Further promising applications of FL are in the areas, such as privacy-preserving driver behavior modeling, anomaly detection, and predictive maintenance. With the advent of cloud infrastructure, 6G, V2X technology, and flying cars, the use of FL models is expected to provide significant breakthroughs.

\bibliographystyle{ieeetr}
\bibliography{ITSC.bib}

\vfill

\begin{IEEEbiography}
[{\includegraphics[width=1in,height=1.25in,clip,keepaspectratio]{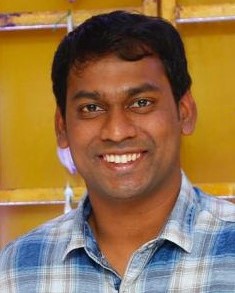}}]
{Vishnu Pandi Chellapandi}
(M'22) received the B.E. degree from the College of Engineering Guindy, Anna University, India, and the M.S. degree from the University of Michigan, Ann Arbor, MI, USA, in 2018. He is currently pursuing the Ph.D. degree with the School of Electrical and Computer Engineering, Purdue University, West Lafayette, IN, USA. He is also working as a Technical Specialist in the Connected and Intelligent Systems group, Cummins Research and Technology, Columbus, IN, USA. His research interests are in the areas of Federated learning, distributed optimization, systems, and optimal controls.
\end{IEEEbiography}

\begin{IEEEbiography}
[{\includegraphics[width=1in,height=1.25in,clip,keepaspectratio]{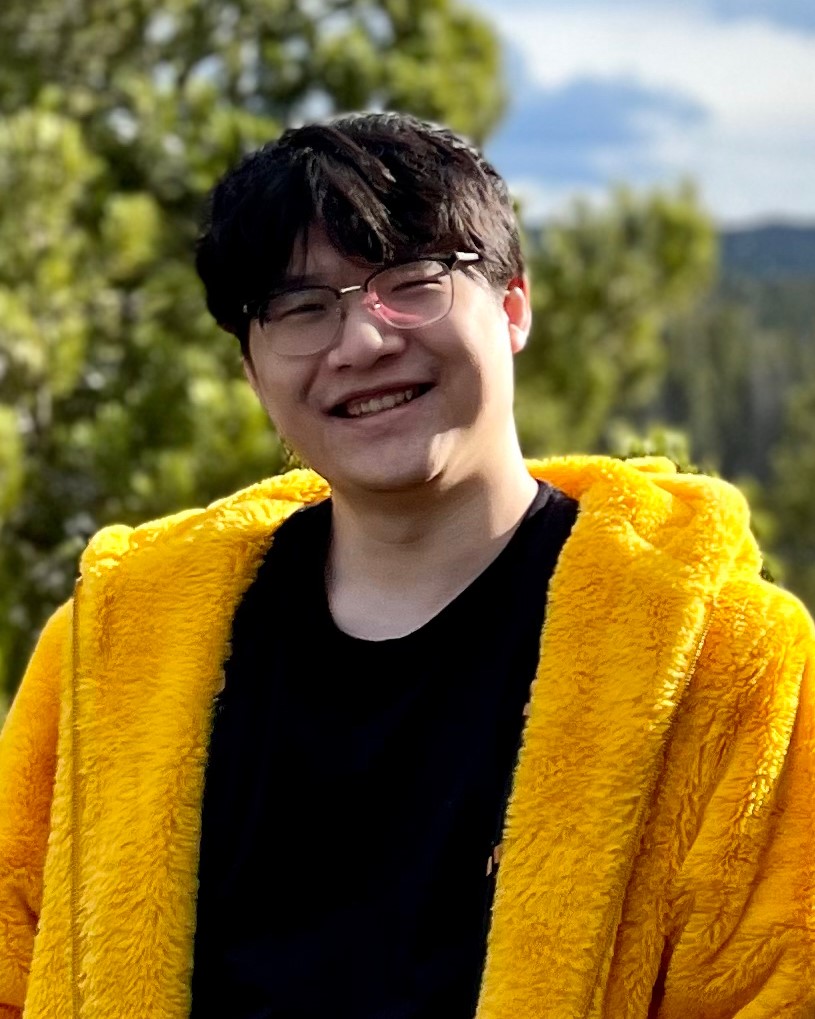}}]
{Liangqi Yuan}
(S'22) received the B.E. degree from the Beijing Information Science and Technology University, Beijing, China, in 2020, and the M.S. degree from the Oakland University, Rochester, MI, USA, in 2022. He is currently pursuing the Ph.D. degree with the School of Electrical and Computer Engineering, Purdue University, West Lafayette, IN, USA. His research interests are in the areas of sensors, the Internet of Things, signal processing, and machine learning.
\end{IEEEbiography}

\begin{IEEEbiography}
[{\includegraphics[width=1in,height=1.25in,clip,keepaspectratio]{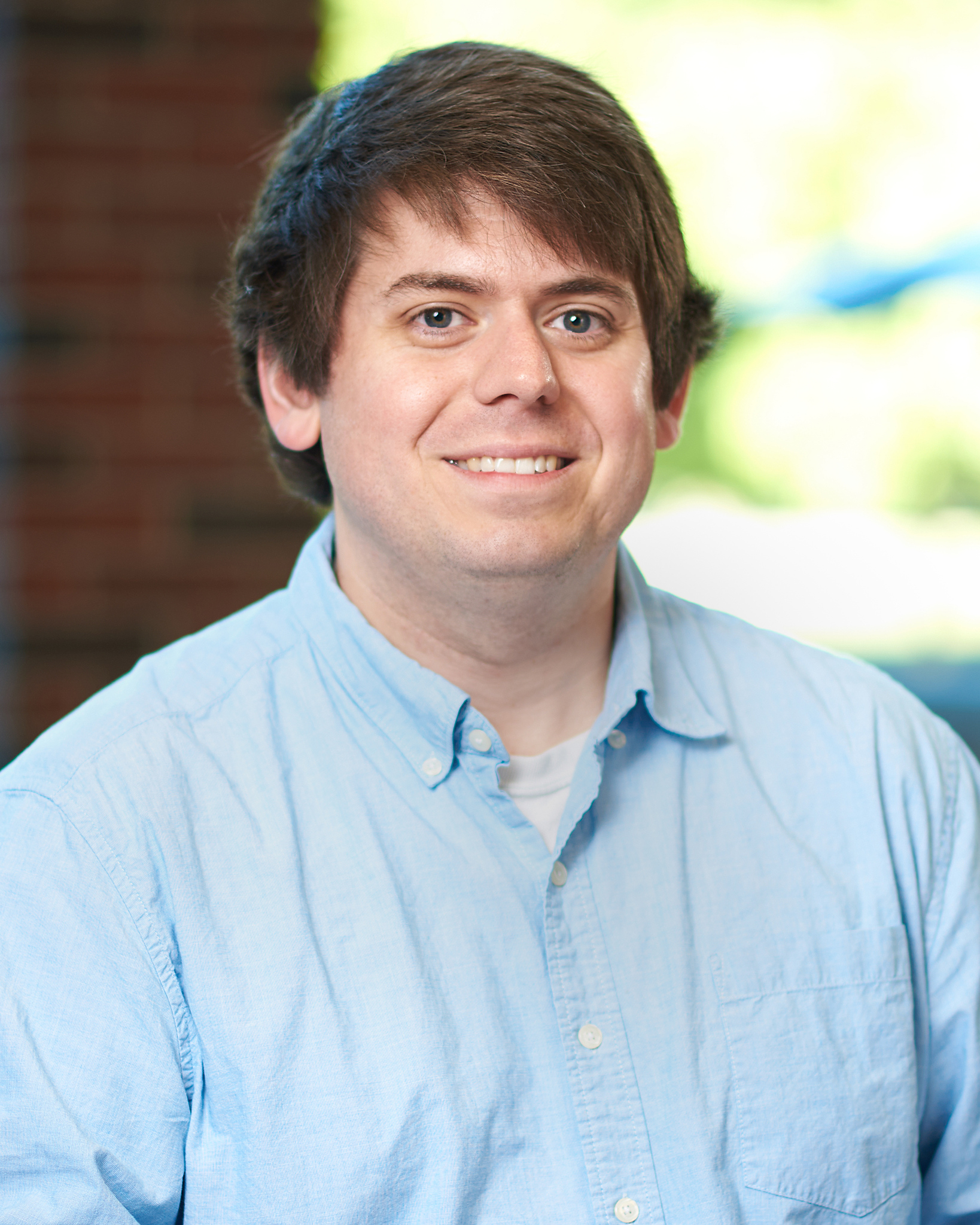}}]
{Christopher G. Brinton}
(S'08, M'16, SM'20) received the M.S. and Ph.D. (with Hons.) degrees in electrical engineering from Princeton University, Princeton, NJ, USA, in 2013 and 2016, respectively. He is the Elmore Assistant Professor with the School of Electrical and Computer Engineering, Purdue University, West Lafayette, IN, USA. Prior to joining Purdue University, he was the Associate Director of the EDGE Lab and a Lecturer of Electrical Engineering with Princeton University. His research interest is at the intersection of networked systems and machine learning, specifically in distributed machine learning, fog/edge network intelligence, and data-driven network optimization. Dr. Brinton is a recipient of the NSF CAREER Award, the ONR Young Investigator Program Award, the DARPA Young Faculty Award, and the Intel Rising Star Faculty Award. He currently serves as an Associate Editor for IEEE Transactions on Wireless Communications, in the ML and AI for wireless area.
\end{IEEEbiography}

\begin{IEEEbiography}
[{\includegraphics[width=1in,height=1.25in,clip,keepaspectratio]{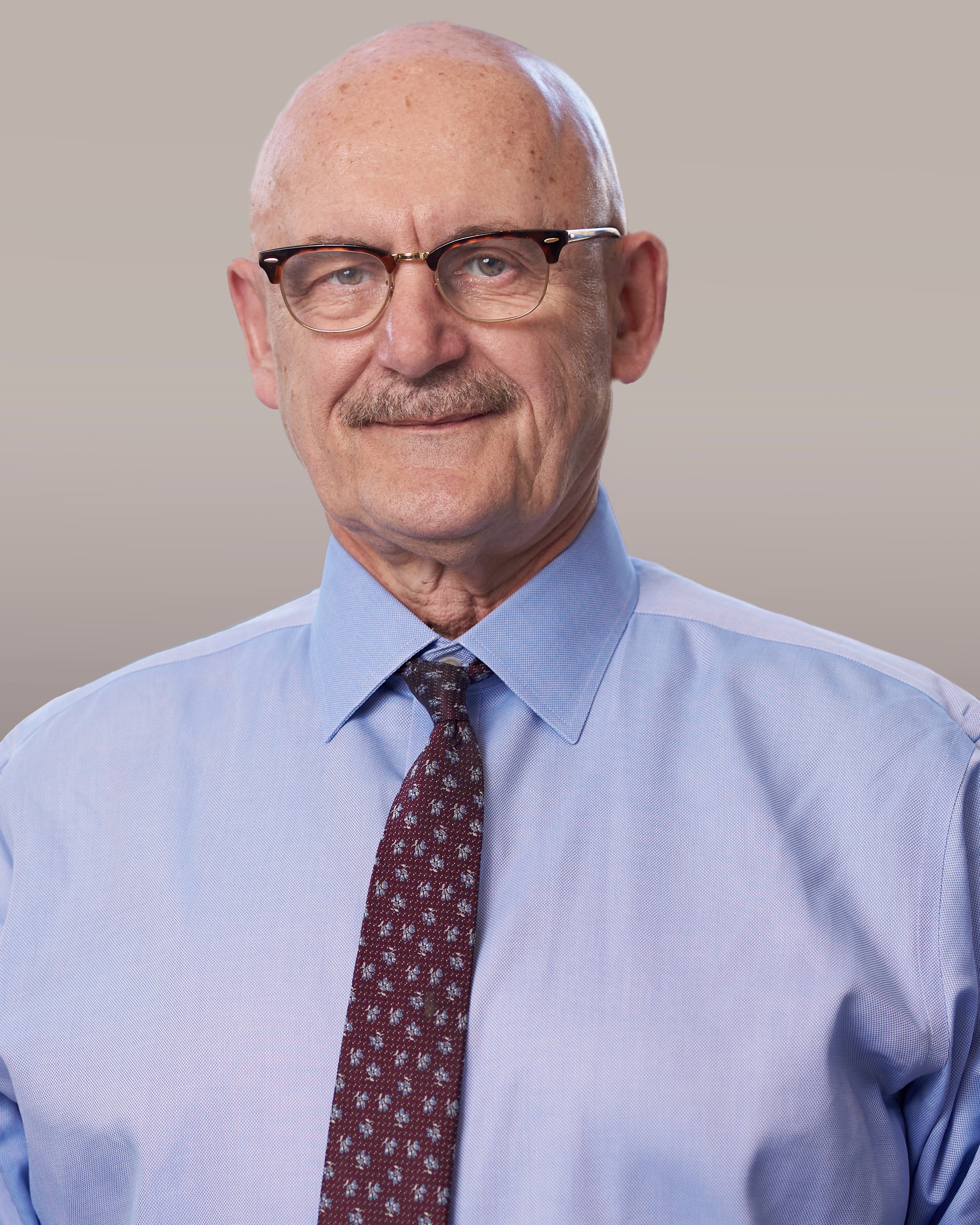}}]
{Stanislaw H. \.{Z}ak}
(LM'21) received the Ph.D. degree in electrical engineering from Warsaw University of Technology, Warsaw, Poland, in 1977. From 1977 to 1980, he was an Assistant Professor with the Institute of Control and Industrial Electronics, Warsaw University of Technology. From 1980 to 1983, he was a Visiting Assistant Professor with the Department of Electrical Engineering, University of Minnesota, Minneapolis, MN, USA. In 1983, he joined the School of Electrical and Computer Engineering, Purdue University, West Lafayette, IN, USA, where he is currently a Professor. He has been involved in various areas of control, optimization, fuzzy systems, and neural networks. He has coauthored {\it Topics in the Analysis of Linear Dynamical Systems} (Polish Scientific Publishers, 1984) and {\it An Introduction to Optimization: With Applications to Machine Learning}, 5th edition (Wiley, 2024), and has authored {\it Systems and Control} (Oxford University Press, 2003). He was an Associate Editor of {\it Dynamics and Control} and the {\it IEEE Transactions on Neural Networks}.
\end{IEEEbiography}

\begin{IEEEbiography}
[{\includegraphics[width=1in,height=1.25in,clip,keepaspectratio]{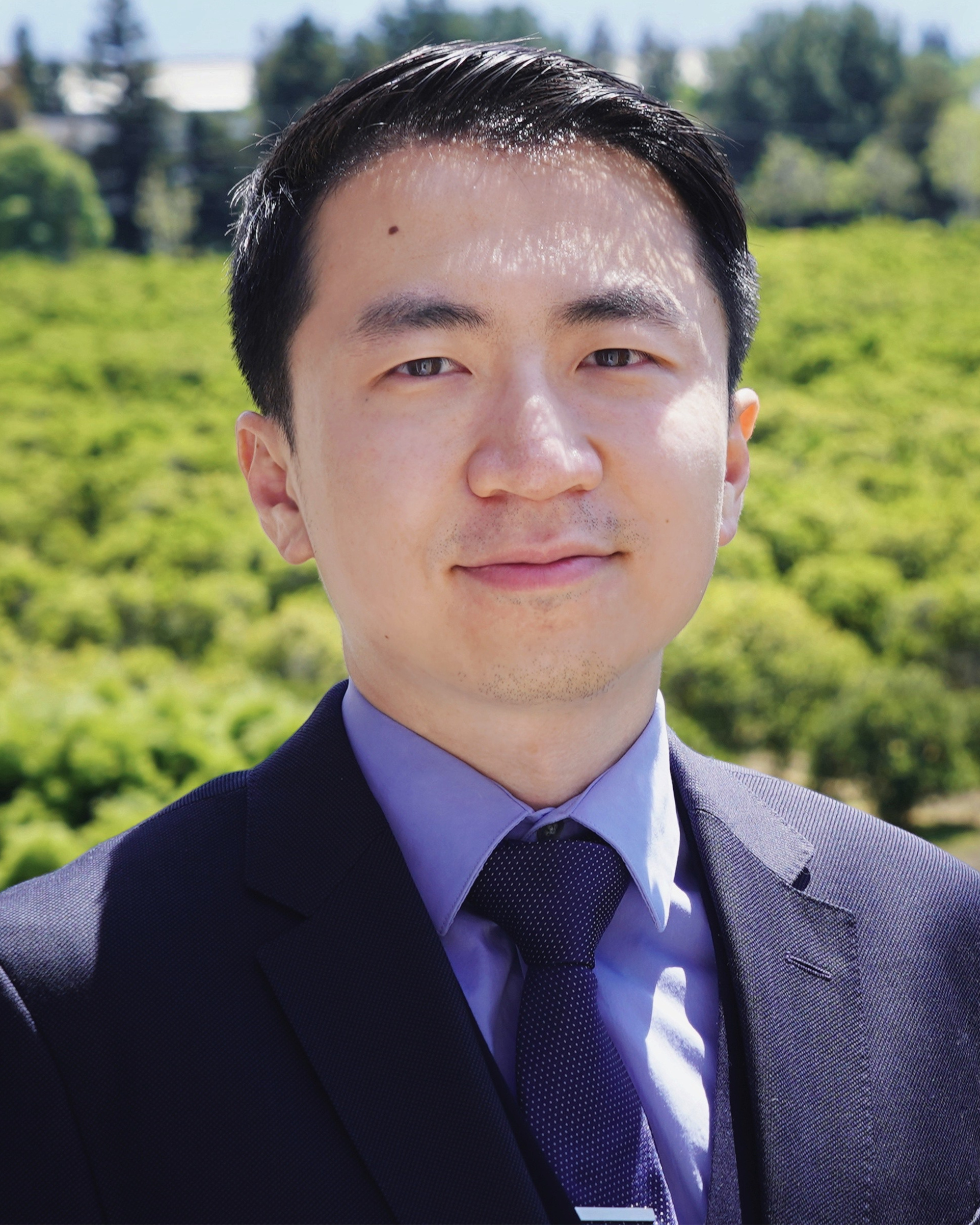}}]
{Ziran Wang}
(S'16-M'19) received the Ph.D. degree in Mechanical Engineering from the University of California, Riverside in 2019. He is a tenure-track assistant professor in the College of Engineering at Purdue University, where he directs the Purdue Digital Twin Lab. Prior to this, Dr. Wang was a principal researcher at Toyota Motor North America R\&D in Mountain View, California. Dr. Wang serves as the founding chair of the IEEE Technical Committee on Internet of Things in Intelligent Transportation Systems, and an associate editor of four journals. His research focuses on automated driving, human-autonomy teaming, and digital twins. 
\end{IEEEbiography}

\end{document}